\RequirePackage{fix-cm}
\documentclass[10pt,journal,compsoc]{IEEEtran}

\usepackage{cuted}
\usepackage[T1]{fontenc}
\usepackage[utf8]{inputenc}
\usepackage{amsmath,amssymb,bm}
\usepackage[final]{microtype}
\usepackage{graphicx,booktabs,array,tabularx,multirow}
\usepackage{url,ragged2e}
\usepackage[table]{xcolor}
\usepackage{tikz,pgfplots}
\usepackage{xcolor}
\usetikzlibrary{arrows.meta,positioning,fit,calc,backgrounds}
\pgfplotsset{compat=1.18}
\usepackage{algorithm,algpseudocode}
\usepackage[nocompress]{cite}
\usepackage{caption,etoolbox}
\renewcommand{\thetable}{\arabic{table}}
\usepackage{stfloats}
\usepackage[nocheckfootnote,nospread]{flushend}
\usepackage{subcaption}
\usepackage{textcomp}
\newcommand{\resultheading}[1]{\par\noindent\textbf{#1.}\enspace}
\definecolor{linkblue}{RGB}{0,65,145}
\definecolor{pcmgray}{RGB}{199,204,210}
\definecolor{pcmblue}{RGB}{45,111,171}
\usepackage[
  pagebackref=true,
  breaklinks=true,
  letterpaper=true,
  colorlinks=true,
  bookmarks=false,
  linkcolor=cyan,
  urlcolor=cyan,
  citecolor=cyan
]{hyperref}
\usepackage{url}
\hypersetup{pdftitle={Motion Mamba 2.0: Autoregressive Diffusion Mamba for Long-Horizon Humanoid Control},
  pdfauthor={Yue Huang, Tianshan Zhang, Zeyu Zhang, Jayeon Kang, Akide Liu, Sungjoon Choi, Dacheng Tao, Hao Tang},
  pdfsubject={Continuous motion generation and humanoid control}}

\graphicspath{{Figures/}}
\newcommand{\method}{Motion Mamba 2.0}
\newcommand{\R}{\mathbb{R}}

\newcommand{\memory}{\mathcal{M}}
\newcommand{\pending}{\widetilde{\mathcal{M}}}
\newcommand{\up}{\ensuremath{\uparrow}}
\newcommand{\dn}{\ensuremath{\downarrow}}
\newcolumntype{L}[1]{>{\raggedright\arraybackslash}p{#1}}
\begin{document}
\title{\textcolor[HTML]{2864B1}{M}%
  \textcolor[HTML]{256AAF}{o}%
  \textcolor[HTML]{2170AD}{t}%
  \textcolor[HTML]{1E76AB}{i}%
  \textcolor[HTML]{1A7CAA}{o}%
  \textcolor[HTML]{1880A6}{n}%
  \ %
  \textcolor[HTML]{1784A1}{M}%
  \textcolor[HTML]{17889C}{a}%
  \textcolor[HTML]{168C97}{m}%
  \textcolor[HTML]{168F90}{b}%
  \textcolor[HTML]{1A9186}{a}%
  \ %
  \textcolor[HTML]{1D937B}{2}%
  \textcolor[HTML]{219571}{.}%
  \textcolor[HTML]{249766}{0}:
   Autoregressive Diffusion Mamba for Long-Horizon Humanoid Control}
\author{Yue Huang$^*$,
        Tianshan Zhang$^*$,
        Zeyu Zhang$^{*\dag}$,
        Jayeon Kang,
        Akide Liu,\\
        Sungjoon Choi,
        Dacheng Tao,~\IEEEmembership{Fellow,~IEEE,}
        Hao Tang$^\ddag$
\IEEEcompsocitemizethanks{
        \IEEEcompsocthanksitem $^*$Equal contribution. $^\dag$Project lead. \protect
        \IEEEcompsocthanksitem $^\ddag$Corresponding author, E-mail: bjdxtanghao@gmail.com. \protect
        \IEEEcompsocthanksitem Yue Huang, Tianshan Zhang, Zeyu Zhang, and Hao Tang are with the School of Computer Science, Peking University, Beijing 100871, China.
        \protect
        \IEEEcompsocthanksitem Jayeon Kang and Sungjoon Choi are with the Department of Artificial Intelligence, Korea University, Seoul 02841, Korea. \protect
        \IEEEcompsocthanksitem Akide Liu is with the Department of Data Science and AI, Monash University, Melbourne VIC 3800, Australia. \protect
        \IEEEcompsocthanksitem Dacheng Tao is with the College of Computing and Data Science, Nanyang Technological University, Singapore. 
    }%
}
\markboth{Submitted to IEEE Transactions on Pattern Analysis and Machine Intelligence}%
{Motion Mamba 2.0: Long-Horizon Humanoid Control}

\IEEEtitleabstractindextext{%
\begin{abstract}\justifying
Long-horizon, real-time controllable motion generation is essential for interactive animation and humanoid robotics, where an agent must respond to changing instructions and spatial goals while moving. Existing autoregressive diffusion methods generate motion incrementally, but bounded attention caches limit access to distant history, while repeated denoising introduces substantial inference overhead. We present \textbf{Motion Mamba 2.0}, a Mamba-based root--body diffusion architecture that integrates local attention, resumable recurrent states, and persistent content memory (PCM) for long-horizon controllable motion generation. PCM transforms compressed motion history from Mamba recurrent states, together with state changes and text-conditioned features, into dynamically updated content slots accessible to local attention. To enable efficient inference, we further develop an execution scheme tailored to this architecture, combining CUDA Graph replay, fused kernels, and execution-flow simplification while preserving the canonical history update. Evaluations on retargeted HiPHI demonstrate accurate terminal control over rollouts of up to 39.52\,s, while short-horizon comparisons on BONES Seed and retargeted HumanML3D show that the model also retains strong spatial control and text--motion matching. On an RTX 4090, the optimized execution reduces median complete-window latency from 1685.35 to 276.14 ms, achieving a median paired speedup of 6.138$\times$ while preserving identical outputs and committed states in the tested rollouts. 
Moreover, we conduct extensive real-world evaluation on the Unitree G1, demonstrating sequential text-guided action composition, expressive motion, and interactive locomotion through keyboard and mouse control.
Code:~\url{https://github.com/AIGeeksGroup/MotionMamba2.0}.
Website:~\url{https://aigeeksgroup.github.io/MotionMamba2.0}.
\end{abstract}
}
\maketitle
\IEEEdisplaynontitleabstractindextext
\IEEEpeerreviewmaketitle
\justifying
\flushbottom

\begin{strip}
\vspace*{-12em} 
\centering
\includegraphics[width=\linewidth]{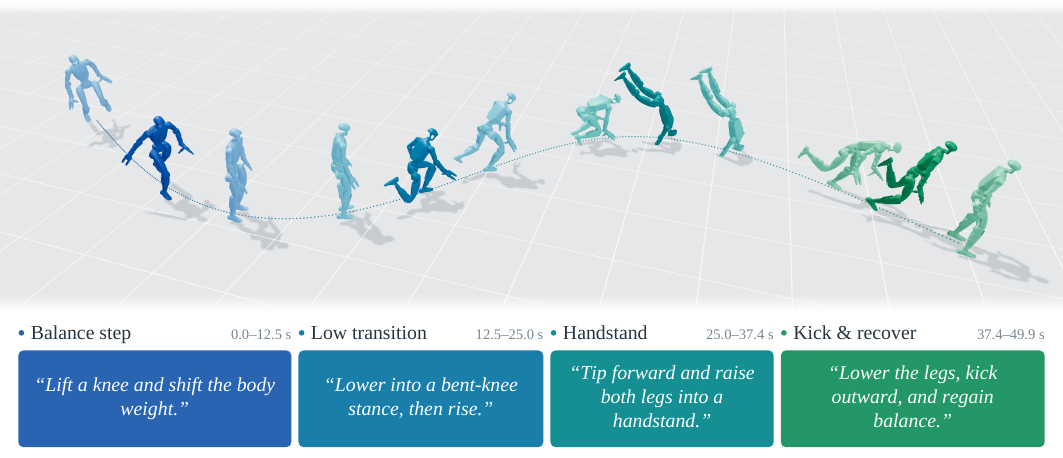}
\vspace{-0.6em} 
\captionof{figure}{\textbf{Motion Mamba 2.0.} Continuous text-guided motion generation for long-horizon humanoid control on the Unitree G1.}
\label{fig:teaser}
\vspace*{1em} 
\end{strip}

\IEEEraisesectionheading{\section{Introduction}\label{sec:introduction}}
\IEEEPARstart{G}{enerating} long-horizon motion from changing instructions is a central problem in interactive animation and humanoid robotics.
A useful generator must interpret an action description, accommodate spatial constraints, and continue from the motion already produced.
These requirements extend beyond synthesizing an isolated clip: a user may change direction during walking, request an upper-body action, or specify a future target while the current motion is still being executed.
The resulting references must be available quickly enough for interactive use and remain compatible with a downstream tracking system.
Unlike offline text-to-motion synthesis, the future instruction sequence need not be known when generation begins.
An interactive generator must therefore preserve useful context while incorporating newly arriving commands, rather than simply replaying a completed trajectory.

Diffusion models produce diverse text-conditioned human motion~\cite{MDM,MotionDiffuse,MLD}.
Recent methods extend this formulation through temporal composition, autoregressive prediction, and direct spatial conditioning~\cite{FlowMDM,AMD,DartControl,ARDY}.
Temporal composition allows multiple descriptions to define a longer sequence, while causal latent models and hierarchical motion tokens support incremental synthesis~\cite{MotionStreamer,MOGO}.
Spatially constrained methods further specify where selected joints or the character should move~\cite{GuidedMotion,OmniControl,COMET,ARDY}.
In particular, generating one window at a time provides a practical interface for online instructions.
However, two difficulties remain closely connected.
A bounded attention cache discards older motion features, whereas an unbounded cache increases storage and computation as the sequence grows.
Moreover, repeated denoising of short windows can incur substantial execution overhead even when each individual model evaluation is small.
An interactive system therefore requires both a suitable representation of motion history and an efficient implementation of its repeated updates.
These requirements concern different aspects of the problem: reducing the number of sampling steps does not extend the history available to the model, and a compact historical state alone does not eliminate denoising overhead.

\begin{figure*}[tb]
\centering
\includegraphics[width=\linewidth,height=0.70\textheight,keepaspectratio]{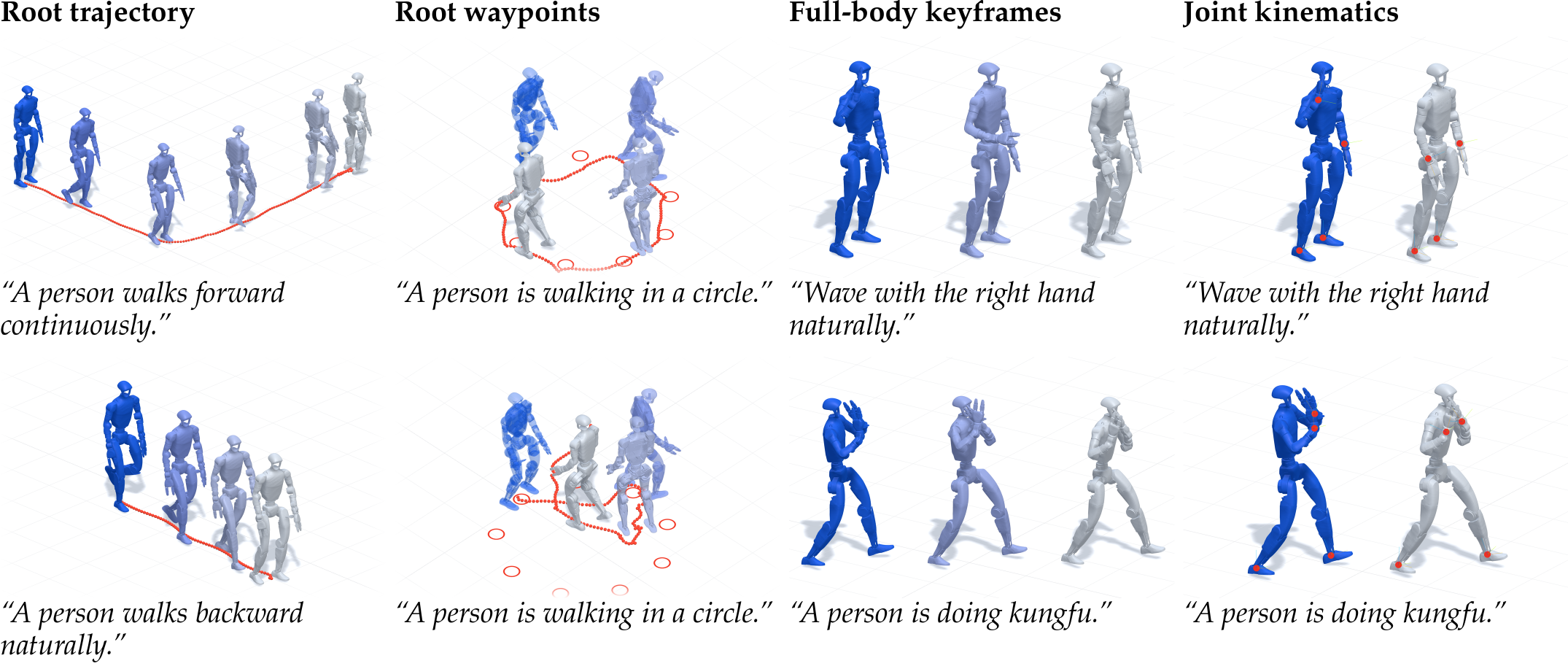}
\caption{\textbf{Interactive motion generation with Motion Mamba 2.0.}
Text instructions specify the requested actions, while spatial controls constrain root motion and selected joint positions. From left to right, the panels show root trajectories, root waypoints, full-body pose sequences, and selected joint positions. Blue-to-gray poses indicate temporal progression, while red overlays visualize trajectories, waypoints, and joint locations. The panels show Viser motion references and a rollout tracked with SONIC.}
\label{fig:introduction}
\end{figure*}

State-space models provide an alternative to storing every historical token by summarizing a sequence in a recurrent state~\cite{Mamba,Mamba3}.
Their use in motion generation has shown the value of efficient temporal modeling~\cite{Motion_Mamba,LightT2M}.
Recent studies also combine Mamba with autoregressive diffusion or use recurrent temporal representations for sustained and interactive motion~\cite{MADM,TCM,DyadicMamba}.
Nevertheless, a compressed recurrent state and an attention cache expose history differently: the former carries evolving context, while the latter permits direct retrieval of explicit features.
Simply placing both branches in a network does not make the recurrent history directly accessible to attention.
This distinction motivates a learned interface between accumulated recurrent motion state and attention-accessible content, rather than relying on either mechanism alone.
Diffusion introduces an additional requirement: successive noisy candidates describe alternatives for the same output window, not successive moments of the accepted motion stream.
Persisting those intermediate candidates would mix sampling iterations with physical time.
We therefore treat the boundary between prediction and history commitment as part of the model's streaming formulation.

We propose \method{}, an autoregressive diffusion model for long-horizon, real-time controllable G1 motion.
The model follows a root--body decomposition~\cite{ARDY}: explicit root predictions determine local trajectory features for body generation.
Within both denoisers, local attention is combined with a resumable Mamba branch.
Our persistent content memory (PCM) learns to map compressed Mamba state summaries, successive state changes, and text-conditioned features into dynamically updated slots accessible to local attention.
The slots complement recent motion features in the bounded cache and aim to support terminal spatial constraint satisfaction and command-conditioned generation over extended rollouts; they are not designed to reproduce arbitrary early instructions verbatim.
Recurrent states, cached features, and slots are committed only after a clean motion window is accepted; intermediate denoising candidates read the same immutable history snapshot.
The fixed-size recurrent state, cache, and PCM slots bound stored context without guaranteeing retention of every detail from arbitrarily distant motion.
The accepted-history separation also guides efficient inference.
Read-only denoising avoids materializing unused state proposals, selective fusion reduces intermediate memory traffic, and CUDA Graphs reduce repeated host dispatch~\cite{CUDAGraphs,Triton}.
These changes preserve root-to-body ordering and the canonical clean-history update.
Our objective is to accelerate the same sampling computation, complementing methods that obtain faster generation through compact representations or altered sampling procedures~\cite{MLD,EMDM,AAMDM}.
We measure their contributions separately, since the effect of local fusion differs from that of graph replay.
The complete scheme generates a 52-frame window in 276.14\,ms at the median on an RTX 4090; this steady-state generation time is distinct from the delay between a user command and its visible execution.

We evaluate the method on three G1 motion benchmarks under a matched 100,000-update training budget (App.~C).
We first examine short- and long-horizon generation on retargeted HiPHI~\cite{HiPHI}, followed by short-horizon comparisons on BONES Seed~\cite{BONES} and retargeted HumanML3D~\cite{HumanML3D}; BONES Seed retains its officially provided G1 trajectories.
The results demonstrate accurate control over extended rollouts together with strong short-horizon performance, while architectural ablations examine the contributions of recurrent state and history-dependent content memory.
Finally, we conduct real-world evaluation on a physical Unitree G1,
which demonstrates sequential text-guided action composition, expressive motion, and interactive locomotion through keyboard and mouse control.

In summary, our contributions are threefold:
\begin{itemize}
\item We propose \textbf{Motion Mamba 2.0}, a Mamba-based autoregressive diffusion model with persistent content memory, connecting accepted recurrent history to local attention for long-horizon, real-time controllable motion generation.
\item We develop and evaluate state-preserving inference optimizations for this architecture, combining execution-flow simplification, fusion kernel, and CUDA Graph replay. On an RTX 4090, the complete optimization achieves a $6.138\times$ median paired speedup.
\item We evaluate the proposed framework on a physical Unitree G1 through multi-stage text-guided action composition and interactive keyboard/mouse control, demonstrating continuous motion execution under changing instructions and spatial commands. The system achieves 94\% full-task success across 100 trials without observed falls.
\end{itemize}

A preliminary version of this work appeared in our ECCV 2024 conference paper, Motion Mamba~\cite{Motion_Mamba}. The present manuscript substantially extends the
conference version in both methodology and experimental evaluation in the following six aspects.
(1) We reformulate the original text-conditioned latent diffusion framework into an autoregressive root--body diffusion architecture for long-horizon humanoid motion generation. Instead of synthesizing
a motion sequence as a single clip, the new framework generates successive 52-frame windows while accommodating changing text instructions and time-indexed spatial constraints.
(2) We redesign the temporal modeling architecture beyond the original Hierarchical Temporal Mamba (HTM) and Bidirectional Spatial Mamba (BSM) blocks. The new denoisers integrate resumable \textbf{Mamba-3 recurrence with local attention and persistent content memory (PCM)}. PCM converts accepted recurrent-state summaries and their changes into gated content slots, providing attention-accessible representations of compressed motion history alongside the sliding cache.
(3) We introduce an accepted-only history update mechanism for autoregressive diffusion. Intermediate noisy predictions read an immutable historical snapshot, whereas the attention cache, recurrent states, and PCM slots are updated together only after a clean motion window is accepted. This separates diffusion sampling iterations from the temporal evolution of generated motion.
(4) We develop a hardware-aware inference pipeline that combines execution-flow simplification, custom fused kernel and CUDA Graph replay across three guidance branches. 
(5) We extend the evaluation from the original
HumanML3D and KIT-ML text-to-motion benchmarks
to a unified G1 motion setting involving retargeted HiPHI, BONES Seed and retargeted HumanML3D. The new protocol evaluates spatial controllability, instruction matching, motion quality, and inference efficiency over autoregressive rollouts of up to 39.52\,s.
(6) We further extend our evaluation to physics-based simulation and real-world deployment \textbf{on a Unitree G1 humanoid}, assessing multi-stage text-guided actions alongside interactive spatial control via keyboard
and mouse.

\section{Related Work}
\label{sec:related_work}
\subsection{Efficient Long-Horizon Motion Generation}
\label{subsec:Efficient_Long-Horizon_Motion_Generation}

Text-conditioned diffusion models learn expressive motion distributions from paired language and motion data~\cite{MDM,MotionDiffuse,MLD}.
Their representations and sampling procedures determine both the cost of synthesis and how motion can be extended in time.
Discrete motion representations offer a complementary formulation, with autoregressive token prediction in T2M-GPT~\cite{T2MGPT} and masked token completion in MMM~\cite{MMM}.
Longer sequences can be constructed by composing action-conditioned segments, extending textual context, or refining overlapping motion intervals~\cite{MultiAct,T2LM,PriorMDM,FlowMDM,STMC,M2D2M}.
Past-conditioned diffusion and temporal motion fields further address continuity beyond isolated clips~\cite{PCMDM,InfiniDreamer,LagrangianMotionFields}.
In particular, Lagrangian Motion Fields represent longer sequences through compact supermotions, whereas FlowMDM blends positional encodings to improve transitions between described segments~\cite{LagrangianMotionFields,FlowMDM}.
These approaches make temporal composition more flexible, but an interactive application additionally requires new conditions to affect a stream that has already begun.

Autoregressive diffusion and accelerated sampling support incremental synthesis with lower per-update cost~\cite{AMD,AAMDM,AMDM,EMDM,DartControl}.
Recent streaming methods introduce causal representations, historical context, or spatial conditioning to accommodate continuous generation~\cite{MotionStreamer,ARDY,FD2,CMDM}.
MOGO instead uses a hierarchical causal Transformer over quantized motion tokens for real-time, extended generation~\cite{MOGO}.
These developments show that motion representation, temporal factorization, and sampling efficiency offer complementary routes to reducing generation cost.

Controllability introduces a further requirement: trajectory and joint targets must influence future motion while preserving the current action.
Guided Motion Diffusion incorporates spatial constraints through representation design and diffusion guidance~\cite{GuidedMotion}, while OmniControl combines spatial and realism guidance to constrain different joints at different times~\cite{OmniControl}.
COMET studies extended joint targets for long-term generation, while ARDY exposes text and flexible kinematic constraints through a hybrid representation~\cite{COMET,ARDY}.
We build on ARDY’s explicit-root and latent-body representation to study how recurrent state, bounded attention, and persistent content can be integrated into its denoising process. The accompanying execution optimizations target repeated inference while preserving the architecture’s sampling and history-update semantics. For a controlled comparison, all methods use independently trained G1 adaptations with the same 100,000-update budget; none uses its original pretrained release.

\subsection{State-Space Models for Motion}
\label{subsec:State_Space_Model}

Structured state-space models connect recurrent, convolutional, and continuous-time sequence processing~\cite{LSSL,S4,DSS,S4D}.
Mamba introduces input-dependent state updates, and subsequent models extend selective scanning to visual and multidimensional data~\cite{Mamba,Vim,VMamba,Mamba-ND}.
Mamba-2 connects structured state-space models with attention through state-space duality~\cite{Mamba2}.
Mamba-3 improves the underlying sequence model through state-space discretization and rotary dynamics~\cite{Mamba3}.
We use this scan as a temporal modeling component; its recurrence is not a contribution of this work.

In motion generation, state-space models have been applied to temporal and spatial dependencies, hybrid architectures, and text-conditioned synthesis~\cite{Motion_Mamba,LightT2M,MotionHMT,TCM}.
Memory-oriented variants also consider longer sequences and the retention of earlier context~\cite{KMM,zhang2024infinimotion}.
MADM combines Mamba with autoregressive diffusion for text-to-motion generation, while Dyadic Mamba and InterMamba extend state-space modeling to interactions between people~\cite{MADM,DyadicMamba,InterMamba}.

Cross-segment memory has also been studied in sequence models: Transformer-XL reuses earlier hidden representations through segment-level recurrence~\cite{TransformerXL}, while Recurrent Memory Transformer propagates dedicated memory tokens between segments~\cite{RMT}.
Our focus is the interface between compressed recurrent motion state and attention during autoregressive diffusion, where intermediate noisy candidates and accepted motion history require separate update rules.
PCM converts accepted state summaries, state changes, and text conditions into dynamically updated content slots, while a bounded attention cache preserves recent motion features.
These slots evolve with accepted motion rather than remaining static, providing dynamically updated motion context without explicitly storing arbitrary past tasks for exact recall.

\subsection{Humanoid Motion Tracking and Control}
\label{subsec:Humanoid_Control}

Humanoid control combines motion planning with the physical constraints of balance, actuation, and contact~\cite{Humanoid_Survey}.
In physics-based character animation, DeepMimic learns control policies that imitate example motions while pursuing task objectives~\cite{DeepMimic}.
Learning-based systems have advanced whole-body teleoperation, human-motion imitation, and expressive tracking~\cite{OmniH2O,HumanPlus,ExBody,MHC}.
Retargeting methods such as GMR~\cite{gmr} align human recordings with a robot embodiment, but kinematic agreement alone does not establish dynamic feasibility.
Large-scale trackers such as SONIC~\cite{SONIC} provide an execution layer for diverse motion references.
This distinction between kinematic generation and physical tracking is important: accurate reference poses do not by themselves guarantee balance or appropriate foot contact after execution.

Motion generation and physical tracking can also be coupled within a control framework.
CLoSD uses a diffusion planner with simulated feedback for multi-task character control~\cite{DiP}, while OMG studies multimodal motion generation for humanoids~\cite{OMG}.
In contrast, our work focuses on generating continuous kinematic references from changing instructions and constraints. These references are executed by an existing tracking system, allowing us to evaluate generation accuracy and inference cost separately from the qualitative simulation and hardware demonstrations of physical execution.

\begin{figure*}[!t]
\centering
\includegraphics[width=\linewidth,height=0.70\textheight,keepaspectratio]{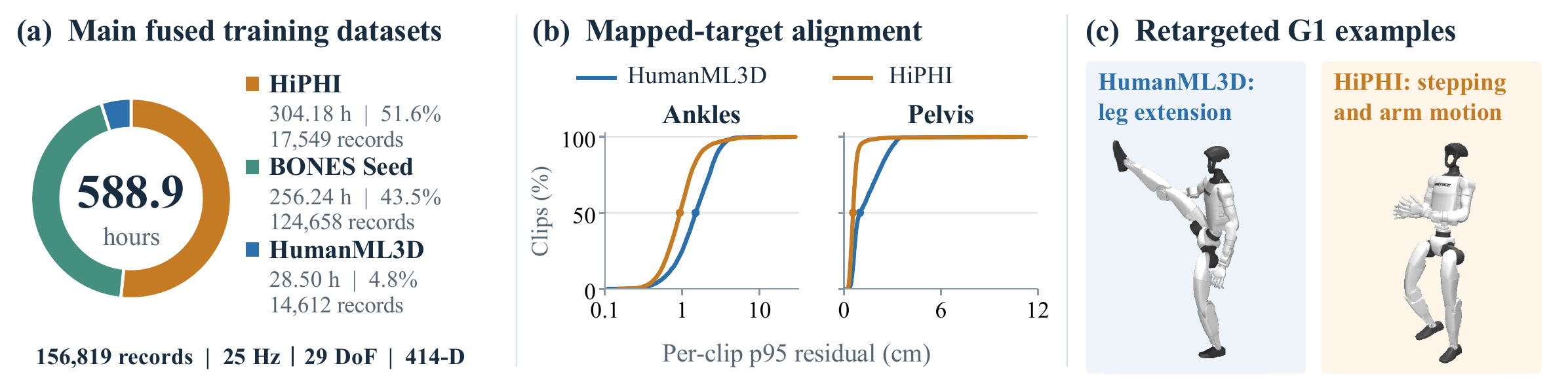}
\caption{\textbf{Fused G1 Datasets and Retargeting Assessment.} (a) Training-set composition by stored duration, including mirrors. (b) Ankle (left, logarithmic axis) and pelvis (right) mapped-target residual distributions over all 38,028 admitted human-source records; dots mark cross-clip medians. (c) Selected poses replayed from stored G1 trajectories. Metric definitions appear in App.~A.}
\label{fig:fusion_retargeting}
\end{figure*}

\section{Datasets and Retargeting}
\label{sec:dataset}

Learning a common motion generator from human and robot recordings requires a consistent embodiment, temporal resolution, and feature representation.
We combine BONES Seed~\cite{BONES}, HumanML3D~\cite{HumanML3D}, and HiPHI~\cite{HiPHI} in a 29-DoF G1 representation with 414 physical features at 25\,Hz.
The resulting training pool contains 156,819 motion records and 588.92 stored hours, including admitted mirrored records.
Fig.~\ref{fig:fusion_retargeting} summarizes the corpus, quantifies lower-body target alignment, and illustrates stored retargeted motions.

\subsection{Dataset Construction}
\label{sec:dataset-construction}

BONES Seed provides robot motion references, while HumanML3D and HiPHI contribute human motion recordings and their associated descriptions.
For HiPHI, we select only the high-precision human-motion subset and exclude the object-interaction subset.
We retarget the selected human body motions to G1; object trajectories and object-interaction supervision are not part of the resulting motion-generation dataset.
The training pool comprises 124,658 BONES Seed records, 14,612 HumanML3D records, and 17,549 HiPHI records.
We retain the source captions, mirror identities, and transformation provenance with each exported record.
The validation and test manifests contain 9,998 and 12,831 additional records, respectively, and are excluded from the training pool.
Counts include admitted mirrors; duration measures stored frames rather than independent capture time.
The complete split statistics are provided in App.~A.

\subsection{Motion Retargeting}
\label{sec:dataset-retargeting}

BONES Seed motions retain their official G1 trajectories through resampling and representation conversion.
For HumanML3D and HiPHI, we use source-specific adapters built on General Motion Retargeting (GMR)~\cite{gmr}.
HumanML3D motions are reconstructed from AMASS~\cite{AMASS} parameters using the SMPL-H body-and-hand model~\cite{SMPLH}, whereas HiPHI motions are reconstructed using BVH forward kinematics.
Rest-pose transforms and pelvis-height scaling align the human skeletons with the common robot asset.
Root translations and local rotations are resampled to 25\,Hz before retargeting.
The solver combines position and orientation objectives under joint limits, followed by temporal refinement and bounded ground and rate corrections.
App.~A details the asset checks and conversion procedure.

\subsection{Retargeting Assessment}
\label{sec:dataset-assessment}

We assess mapped-target alignment over all 18,525 admitted HumanML3D and 19,503 admitted HiPHI records across the training, validation, and test splits.
For each clip, we compute the 95th-percentile distance between a robot body origin and its scaled, offset human target; paired ankles use the worse side, and clips receive equal weight.
Fig.~\ref{fig:fusion_retargeting}(b) shows the ankle and pelvis distributions for the complete admitted cohorts, without selecting low-error clips.
The cross-clip median ankle residuals are 1.49\,cm on HumanML3D and 0.93\,cm on HiPHI; the corresponding cross-clip 95th percentiles are 3.37 and 2.39\,cm.
Median pelvis residuals are 0.99 and 0.56\,cm, respectively.
These statistics quantify alignment to the mapped lower-body objectives.

The stored examples in Fig.~\ref{fig:fusion_retargeting}(c) illustrate leg extension, stepping, and coordinated arm motion.
Such pose-level evidence and mapped-target position residuals describe different aspects of retargeting: a target offset alone does not establish loss of the depicted action, and recognizable poses do not establish precise hand contact.
Separately, conversion between the stored motion formats and the 414-D representation has a maximum round-trip position discrepancy of $1.31\times10^{-6}$\,m over the admitted human-source records.

\section{The Proposed Method}
\label{sec:method}

\begin{figure*}[!t]
  \centering
  \includegraphics[width=\linewidth,height=0.70\textheight,keepaspectratio]{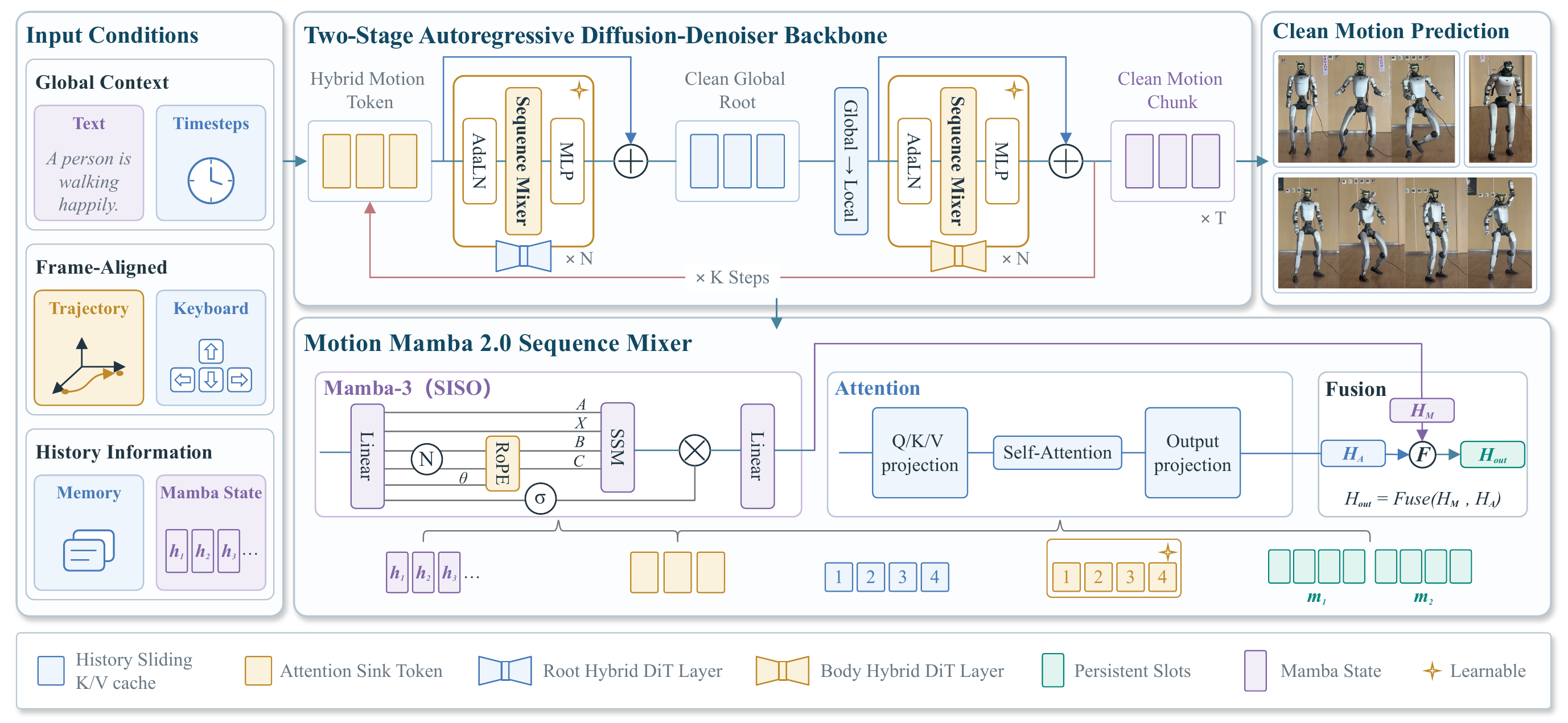}
  \caption{\textbf{Overview of \method{}.} Conditioned on text, spatial controls, and accepted motion history, the framework generates motion autoregressively through a two-stage diffusion denoiser. At each denoising step, the root network predicts the global trajectory, whose local features condition body prediction. Both networks combine attention and Mamba-3 through a hybrid sequence mixer. After each clean motion window is accepted, the history states are updated for subsequent generation.}
  \label{fig:overview}
\end{figure*}

As shown in Fig.~\ref{fig:overview}, \method{}
combines autoregressive root--body diffusion with
local attention, resumable Mamba-3 states, and
persistent content memory (PCM). Local attention
retrieves recent motion, while PCM supplies attention-readable content derived from the accumulated recurrent state.
The local cache remains bounded, while Mamba states and PCM slots carry forward information from accepted motion windows. All historical states are frozen during denoising and committed only after a clean motion window is accepted.

\subsection{Streaming Formulation and Motion Representation}
\label{sec:preliminary}
\label{sec:representation}

We divide the output into windows of $C$ tokens, with each token representing $P$ consecutive frames. At window $t$, the conditions comprise a text instruction $s_t$, spatial targets $\mathbf{g}_t$, and their validity mask $\mathbf{v}_t$. The memory $\memory_{t-1}$ is constructed from the accepted motion prefix. The conditional distribution is factorized over windows:
\begin{equation}
 p_\theta(\mathbf{x}_{0:T}\mid\mathbf{s},\mathbf{g},\mathbf{v})
 =\prod_{t=0}^{T}p_\theta
 \bigl(\mathbf{x}_t\mid s_t,\mathbf{g}_t,\mathbf{v}_t;\memory_{t-1}\bigr).
 \label{eq:factorization}
\end{equation}
Text specifies the desired behavior, whereas spatial conditions identify selected quantities and their target times. Targets may lie within or beyond the current output window. They provide conditioning information without requiring the model to emit the intervening future motion in advance.

We adopt the hybrid representation of ARDY~\cite{ARDY}, retaining an explicit root trajectory and compressing articulation with a causal motion codec. At each frame, the root comprises position $\mathbf{p}\in\R^3$ and heading $\psi$. Body features include joint rotations, joint positions and velocities, and contact indicators. A token concatenates a root patch $\mathbf{r}_i\in\R^{5P}$ with a body latent $\mathbf{b}_i\in\R^L$:
\begin{equation}
 \mathbf{m}_{r}=[\mathbf{p};\cos\psi;\sin\psi],
 \quad \mathbf{x}_i=[\mathbf{r}_i;\mathbf{b}_i].
 \label{eq:hybrid}
\end{equation}
The encoder $\mathcal{E}$ maps explicit motion $\mathbf{m}$ to body latents. The decoder $\mathcal{D}$ reconstructs articulation from these latents and local root features computed by $\operatorname{Local}(\cdot)$:
\begin{equation}
 \mathbf{b}_{1:C}=\mathcal{E}(\mathbf{m}_{1:CP}),
 \quad\widehat{\mathbf{m}}_{b,1:CP}
 =\mathcal{D}\bigl(\mathbf{b}_{1:C},
 \operatorname{Local}(\mathbf{r}_{1:C})\bigr).
\label{eq:tokenizer}
\end{equation}
This representation exposes the trajectory directly to spatial control while reducing the dimensionality of articulated motion. Coordinates are expressed in the current canonical frame; the accumulated translation is retained separately to recover the world-space trajectory.

We use $\tau$ for diffusion time and $\ell$ for the layer index.
Following the forward diffusion process~\cite{DDPM}, a clean
hybrid window $\mathbf{x}_t$ is perturbed with Gaussian noise
according to a predefined diffusion noise schedule:
\begin{equation}
 \mathbf{z}_{t,\tau}
 =\sqrt{\bar\alpha_\tau}\,\mathbf{x}_t
 +\sqrt{1-\bar\alpha_\tau}\,\boldsymbol\epsilon,
 \quad \boldsymbol\epsilon\sim\mathcal{N}(\mathbf{0},\mathbf{I}).
 \label{eq:forward}
\end{equation}
Here, $\bar\alpha_\tau=\prod_{s=1}^{\tau}(1-\beta_s)$,
where $\beta_s$ is the noise variance at diffusion step $s$.
The denoiser predicts clean hybrid features.
A fixed sampling schedule repeatedly refines the noisy window
while its accepted-history memory remains unchanged.

\subsection{Recurrent Root--Body Denoising}
\label{sec:denoiser}

\noindent\textbf{Two-stage prediction.}
Following the root--body decomposition, each diffusion evaluation first predicts the global root and then predicts articulation conditioned on that root. The conditioning vector $\mathbf{c}_{t,\tau}$ combines text, diffusion time, and the available initial heading. The root input $\mathbf{u}^{r}_{t,\tau}$ contains noisy hybrid features and temporally aligned controls. Valid root observations replace the corresponding noisy coordinates; body observations and their masks are supplied as aligned features. Observed future targets are embedded at their target positions, and empty future entries are masked out. The body input $\mathbf{u}^{b}_{t,\tau}$ concatenates local features of the predicted root, noisy body latents, and aligned observations. The root and body networks are denoted by $\mathcal{F}^r_\theta$ and $\mathcal{F}^b_\theta$:
\begin{equation}
\begin{aligned}
 \widehat{\mathbf{r}}_{t,0}
 &=\mathcal{F}^{r}_\theta
(\mathbf{u}^{r}_{t,\tau},\mathbf{c}_{t,\tau};\memory^r_{t-1}),\\
 \widehat{\mathbf{b}}_{t,0}
 &=\mathcal{F}^{b}_\theta
 \bigl(\mathbf{u}^{b}_{t,\tau},
 \mathbf{c}_{t,\tau};\memory^b_{t-1}\bigr).
\end{aligned}
\label{eq:root_body}
\end{equation}
The sampler consumes the concatenated clean prediction $[\widehat{\mathbf{r}}_{t,0};\widehat{\mathbf{b}}_{t,0}]$. Thus, root prediction is refreshed before body prediction at every diffusion step. The two stacks have separate parameters and layer-specific memories.
This decomposition exposes global translation and heading for direct spatial conditioning, while allowing articulated motion to adapt to the predicted root trajectory. By updating the root-conditioned body input at each denoising step, the two predictions remain coupled throughout diffusion sampling rather than being combined only after generation.

\noindent\textbf{Hybrid sequence layers.}
Each stack consists of diffusion transformer layers~\cite{DiT} with adaptive normalization and a gated residual multilayer perceptron. Inside the sequence mixer, attention and Mamba process the same normalized, condition-modulated features $\mathbf{U}^\ell_{t,\tau}$. Attention retrieves recent features, whereas Mamba resumes its state from the accepted prefix. Their outputs are combined by a learned layerwise gate. PCM augments the attention path as described in Sec.~\ref{sec:memory} and it does not replace the recurrent branch.

\noindent\textbf{Mamba-3 recurrence.} We adopt the resumable SISO Mamba-3 scan~\cite{Mamba3}.
For one head, let $\mathbf{u}_i$ denote the value
projection, $\mathbf{B}_i$ and $\mathbf{C}_i$ the
normalized state-input and readout vectors,
$A_i<0$ the decay, $\Delta_i>0$ the step size,
$\lambda_i\in(0,1)$ the trapezoidal coefficient,
and $\boldsymbol\theta_i$ the phase parameter.
The operator $\mathcal{R}$ rotates paired state
coordinates. Omitting window, layer, and diffusion
indices, the recurrence is
\begin{equation}
\begin{aligned}
 \boldsymbol\phi_i
 &=\boldsymbol\phi_{i-1}+\pi\Delta_i\tanh(\boldsymbol\theta_i),\\
 \widetilde{\mathbf{B}}_i
 &=\mathcal{R}(\boldsymbol\phi_i)\mathbf{B}_i,
 \quad
 \widetilde{\mathbf{C}}_i
 =\mathcal{R}(\boldsymbol\phi_i)\mathbf{C}_i,\\
 \alpha_i&=\exp(\Delta_i A_i),
 \quad\beta_i=(1-\lambda_i)\Delta_i\alpha_i,
 \quad\gamma_i=\lambda_i\Delta_i,\\
 \mathbf{H}_i
 &=\alpha_i\mathbf{H}_{i-1}
 +\beta_i\mathbf{u}_{i-1}\widetilde{\mathbf{B}}_{i-1}^{\top}+\gamma_i\mathbf{u}_i\widetilde{\mathbf{B}}_i^{\top},\\
\mathbf{y}_i&=\mathbf{H}_i\widetilde{\mathbf{C}}_i+D\mathbf{u}_i.
\end{aligned}
\label{eq:mamba_recurrence}
\end{equation}
An output gate and projection map the head outputs back to the layer width. Eq.~\eqref{eq:mamba_recurrence} combines adjacent state inputs through the trapezoidal update and represents relative phase through accumulated rotations. It provides a mechanism for retaining evolving dynamics; its contribution to motion generation is assessed empirically. The carried state includes the recurrent matrix, accumulated phase, and previous state-input and value vectors required to resume the scan. Its size is independent of the accepted sequence length.

\subsection{Persistent Content Memory}
\label{sec:memory}

\begin{figure*}[!t]
    \centering
    \includegraphics[width=\linewidth]{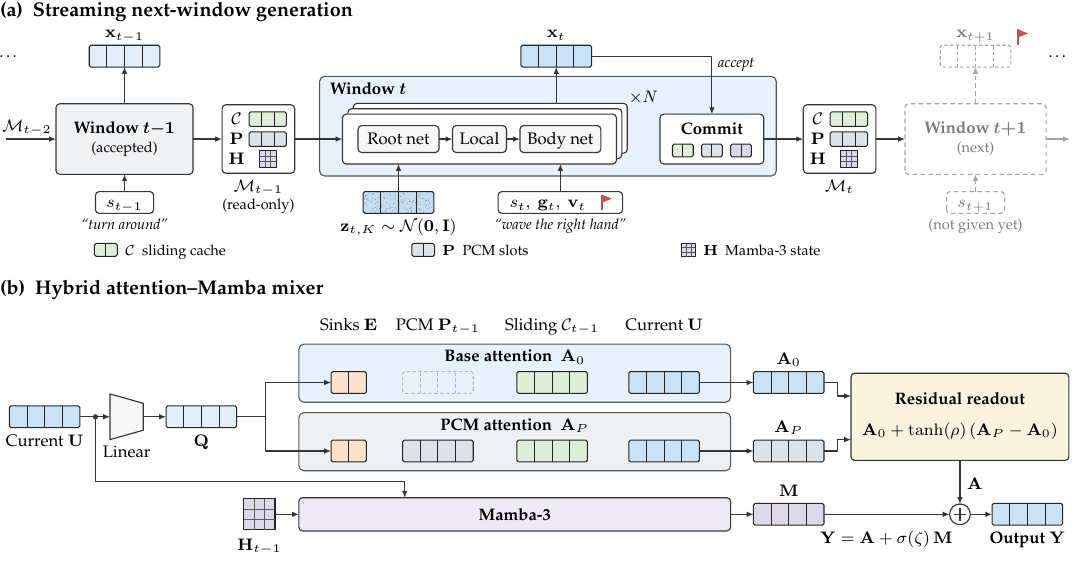}
    \caption{\textbf{Streaming generation and hybrid attention--Mamba mixing.}
    \textbf{(a)} Motion is generated window by window through root--body
    diffusion, with the history held fixed during denoising and updated
    only after the sampled window is accepted.
    \textbf{(b)} Base attention retrieves recent motion features, while
    PCM attention additionally accesses persistent content slots.
    Their gated residual difference is combined with the Mamba-3
    output resumed from the committed recurrent state.}
    \label{fig:hybrid}
\end{figure*}

The recurrent state and attention cache play complementary roles. The bounded cache directly exposes recent keys and values but evicts older features; Mamba carries a fixed-size compressed representation of accepted motion history, which is not itself an attention token. PCM learns a state-to-attention mapping from recurrent summaries, their changes, and text conditions to content slots. These slots make the accumulated motion context available to subsequent windows without claiming exact recovery of arbitrary evicted events.

\noindent\textbf{Local retrieval and static sinks.}
In each layer, the cache $\mathcal{C}^\ell_{t-1}$ stores projected keys, values, and validity masks from the latest $W$ accepted windows. Let $\mathbf{E}^\ell$ be learned sink tokens, $\mathbf{U}^\ell$ the current noisy features, and $\mathbf{Q}^\ell$ their learned query projection. The functions $K^\ell(\cdot)$ and $V^\ell(\cdot)$ are shared by sinks, current tokens, and PCM slots. Historical keys and values are denoted by $\mathbf{K}^\ell_h$ and $\mathbf{V}^\ell_h$. The base attention operator $\operatorname{Attn}^\ell$ includes the output projection and token-validity mask:
\begin{equation}
\begin{aligned}
 \mathbf{K}^\ell_0
 &=[K^\ell(\mathbf{E}^\ell);\mathbf{K}^\ell_h;K^\ell(\mathbf{U}^\ell)],\\
 \mathbf{V}^\ell_0
 &=[V^\ell(\mathbf{E}^\ell);\mathbf{V}^\ell_h;V^\ell(\mathbf{U}^\ell)],\\
 \mathbf{A}^\ell_0
 &=\operatorname{Attn}^\ell
 (\mathbf{Q}^\ell,\mathbf{K}^\ell_0,\mathbf{V}^\ell_0).
\end{aligned}
\label{eq:attention}
\end{equation}
Current-window attention is bidirectional because the noisy window is denoised jointly. The sinks are static model parameters, distinct from history-dependent PCM slots. They remain available when the cache is short or empty; they do not store an episode's past motion.

\noindent\textbf{Content construction and gated writing.}
At layer $\ell$, PCM maintains $S$ memory slots
$\mathbf{P}_t^\ell\in\R^{S\times d}$, initialized to zero.
The projector $\Pi^\ell$ pools the accepted recurrent matrix,
applies a learned projection, and normalizes the resulting
state summary. We compute this summary and its difference
from the previous accepted window as
\begin{equation}
\label{eq:state_summary_difference}
\overline{\mathbf{s}}_t^\ell
= \Pi^\ell(\mathbf{H}_t^\ell),
\quad
\mathbf{d}_t^\ell
= \overline{\mathbf{s}}_t^\ell
- \overline{\mathbf{s}}_{t-1}^\ell.
\end{equation}
The summary captures accumulated context, while the
difference describes the latest accepted transition.

Let $\mathbf{e}_t$ denote the pooled embedding of the active
text instruction. The write path projects this embedding
into a layer-specific representation and predicts
featurewise modulation parameters:
\begin{equation}
\label{eq:text_write_condition}
\begin{aligned}
\mathbf{e}_t^\ell
= \operatorname{LN}
   (W_e^\ell\mathbf{e}_t+\mathbf{b}_e^\ell),
\quad f_c^\ell(\mathbf{e}_t^\ell)
= [\boldsymbol{\gamma}_t^\ell;
    \boldsymbol{\beta}_t^\ell].
\end{aligned}
\end{equation}
These parameters condition the normalized state difference
on the active instruction:
\begin{equation}
\label{eq:text_modulated_difference}
\widetilde{\mathbf{d}}_t^\ell
= (1+\tanh\boldsymbol{\gamma}_t^\ell)
  \odot\operatorname{LN}(\mathbf{d}_t^\ell)
  +\boldsymbol{\beta}_t^\ell.
\end{equation}

We concatenate the state summary, the text-modulated
difference, and the projected text embedding to construct
the candidate slot content:
\begin{equation}
\label{eq:candidate_content}
\begin{aligned}
\mathbf{j}_t^\ell
&= [\overline{\mathbf{s}}_t^\ell;
    \widetilde{\mathbf{d}}_t^\ell;
    \mathbf{e}_t^\ell],
\quad\mathbf{Z}_t^\ell
= \operatorname{LN}
   \bigl(f_p^\ell(\mathbf{j}_t^\ell)\bigr).
\end{aligned}
\end{equation}
Slot normalization acts along the feature dimension.

Finally, a gate computed from the joint representation and
the previous memory controls the write rate:
\begin{equation}
\label{eq:content_write}
\begin{aligned}
\mathbf{w}_t^\ell
&= \eta\,\sigma\!\left(
   f_w^\ell\!\left(
   [\mathbf{j}_t^\ell;
    \operatorname{vec}(\mathbf{P}_{t-1}^\ell)]
   \right)\right),\\
\mathbf{P}_t^\ell
&= (1-\mathbf{w}_t^\ell)
   \odot\mathbf{P}_{t-1}^\ell
   +\mathbf{w}_t^\ell\odot\mathbf{Z}_t^\ell.
\end{aligned}
\end{equation}
The write-rate cap $\eta\in(0,1]$ bounds each gate value,
limiting the contribution of candidate content to each
update. Text conditions both candidate-content generation
and write gating, allowing the same state transition to
produce different memory updates under different
instructions. The slots therefore accumulate
session-specific content rather than serving as fixed
learned vectors. The text condition excludes diffusion-time
and initial-heading embeddings. Implementation details
appear in App.~B.

\noindent\textbf{Residual attention readout.}
The current window reads $\mathbf{P}^\ell_{t-1}$ without modifying it. Augmented keys $\mathbf{K}^\ell_P$ and values $\mathbf{V}^\ell_P$ insert projected slots immediately after the sinks in Eq.~\eqref{eq:attention}. Base and augmented attention use the same query $\mathbf{Q}^\ell$. A learned scalar $\rho_\ell$ controls the attention difference, and $\zeta_\ell$ controls the Mamba contribution $\mathbf{M}^\ell$:
\begin{equation}
\begin{aligned}
 \mathbf{A}^\ell_P
 &=\operatorname{Attn}^\ell
 (\mathbf{Q}^\ell,\mathbf{K}^\ell_P,\mathbf{V}^\ell_P),\\
 \mathbf{Y}^\ell
 &=\mathbf{A}^\ell_0+\tanh(\rho_\ell)
 (\mathbf{A}^\ell_P-\mathbf{A}^\ell_0)+\sigma(\zeta_\ell)\mathbf{M}^\ell.
\end{aligned}
\label{eq:content_read}
\end{equation}
Initializing $\rho_\ell$ to zero recovers the Native mixer output exactly, despite the additional slots. This differs from initializing slot values to zero, since extra keys can still change attention normalization. PCM leaves query construction unchanged; state summaries enter through content writing rather than through a separate state-dependent query. This formulation targets usable motion context instead of explicit indexing of past instructions.

\noindent\textbf{Accepted-only state updates.}
Each denoising evaluation reads an immutable snapshot. The denoiser $\mathcal{F}_\theta$ may produce a temporary proposal $\pending_{t,\tau}$, but this proposal is not committed during sampling:
\begin{equation}
 (\widehat{\mathbf{x}}_{t,0},\pending_{t,\tau})
 =\mathcal{F}_\theta
 (\mathbf{z}_{t,\tau},\mathbf{c}_{t,\tau};\memory_{t-1}).
 \label{eq:readonly}
\end{equation}
After sampling, the accepted window is recentered and materialized as clean history in a canonical pass at diffusion time zero. This pass updates only accepted motion tokens. Layerwise cache append, recurrent-state replacement, and PCM writing are committed together across the root and body stacks. If the pass fails, both stacks retain their previous state. The streaming generation algorithm in App.~B.2 summarizes the procedure.

The exact cache grows only to $W$ windows, and each recurrent state and slot bank has fixed size. For a fixed output window and conditioning budget, memory and computation per new window therefore do not grow with elapsed generation time. This bound concerns computational context; it does not imply error-free motion over arbitrary durations.

\subsection{State-Preserving Efficient Inference}
\label{sec:efficient_execution}

Repeated diffusion denoising incurs overhead from
kernel launches, intermediate memory traffic, and
unnecessary state materialization. We address these
costs through state-preserving execution optimizations
while maintaining root--body dependencies and
accepted-only history updates.

\noindent\textbf{Read-only execution.}
During denoising, we avoid materializing state proposals and diagnostics that are not consumed by either the prediction or a later commit. All recurrent computation contributing to predicted motion is retained. The canonical clean-history pass still constructs and commits the complete accepted state.

\noindent\textbf{Fusion Kernel.}
We fuse Mamba transition preprocessing, including the decay activation, decay floor, step-size transformation, and product $\Delta_iA_i$, into a single traversal. Intermediate values remain in FP32 registers before outputs are written in the scan's required layout. Compatible scale--shift and gated-residual operations are also fused in the execution variants that pass the numerical checks. The attention operator and recurrent scan are retained. Exact preprocessing and precision boundaries are given in App.~B.

\suppressfloats[t]
\begin{figure}[t]
  \centering
  \includegraphics[width=\linewidth]
    {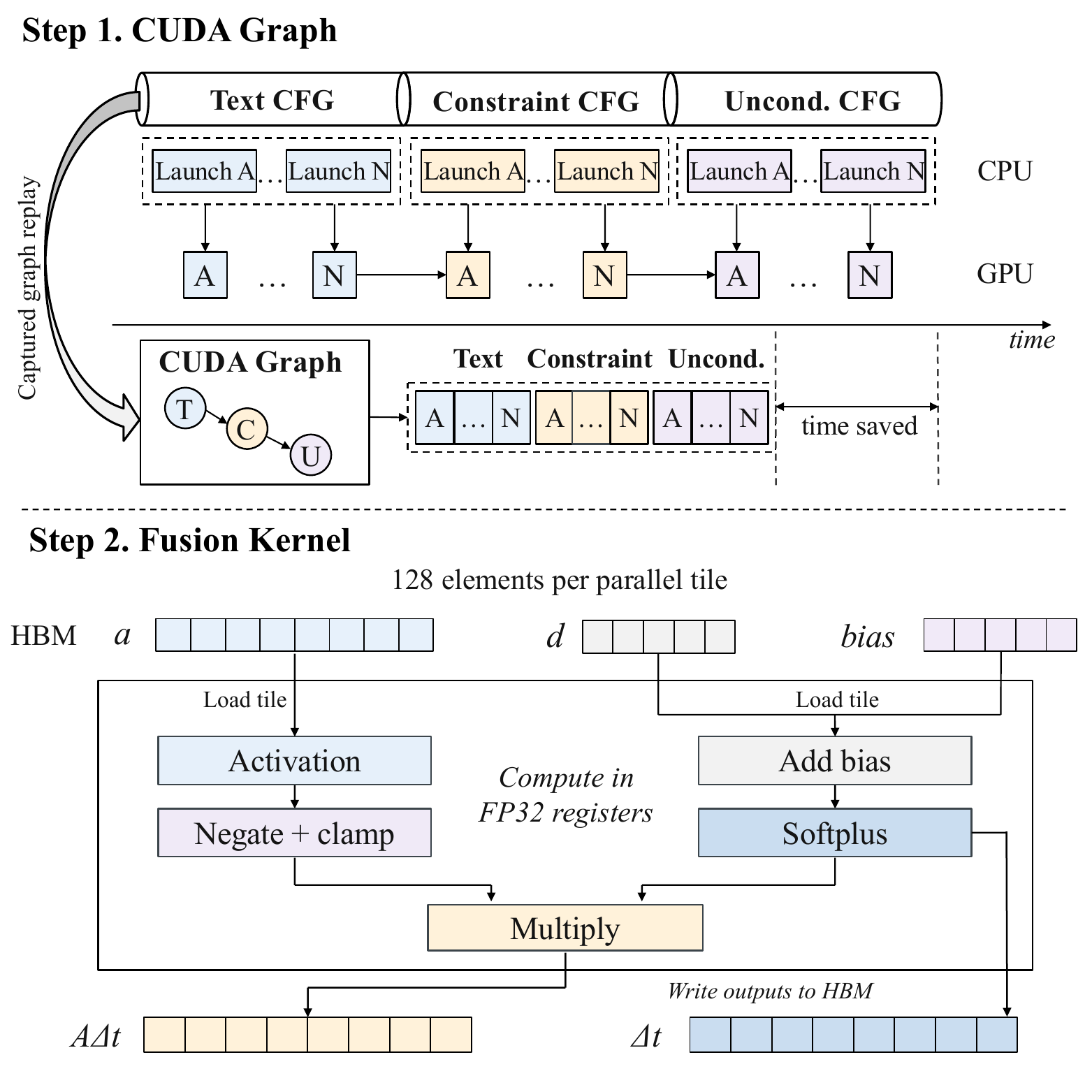}
  \caption{\textbf{CUDA Graph and Kernel Fusion.}
    Stack-level CUDA Graph replay of the three CFG branches,
    with separate graphs for the root and body DiT stacks,
    and fused Mamba transition preprocessing.}
  \label{fig:cuda_graph_fusion}
\end{figure}

\noindent\textbf{CUDA Graph replay.}
The root and body DiT sequence stacks use separate CUDA Graphs. Within each stack, the text-conditioned, spatially conditioned, and unconditional CFG branches execute serially and are captured together. Input projections, conditioning preparation, root-output processing, root-to-local conversion, and body-input construction remain outside these stack-level graphs. Stable input and state buffers are refreshed before replay. Sampling updates, random-number handling, recentering, and accepted-state commits also remain outside capture. Graphs are rebuilt when tensor layouts or parameter versions become incompatible; unsupported calls use the reference path.

\subsection{Learning Objective}
\label{sec:learning}

We optimize clean-window prediction with four terms. The hybrid term $\mathcal{L}_{\mathrm{hyb}}$ applies Smooth L1 to explicit root patches and body latents. The decoded-body term $\mathcal{L}_{\mathrm{body}}$ applies L1 to reconstructed articulation. The goal term $\mathcal{L}_{\mathrm{goal}}$ applies L1 only to observed coordinates inside the generated window; later goals remain conditioning inputs until their target frames enter the output window. The consistency term $\mathcal{L}_{\mathrm{FK}}$ measures Euclidean disagreement between joint positions obtained from decoded rotations through forward kinematics and those represented directly by decoded position features. Each term is normalized over its applicable elements. Their coefficients are denoted by $\lambda_{\mathrm{hyb}}$, $\lambda_{\mathrm{body}}$, $\lambda_{\mathrm{goal}}$, and $\lambda_{\mathrm{FK}}$:
\begin{equation}
\begin{aligned}
 \mathcal{L}
 &=\lambda_{\mathrm{hyb}}\mathcal{L}_{\mathrm{hyb}}
 +\lambda_{\mathrm{body}}\mathcal{L}_{\mathrm{body}}+\lambda_{\mathrm{goal}}\mathcal{L}_{\mathrm{goal}}
 +\lambda_{\mathrm{FK}}\mathcal{L}_{\mathrm{FK}}.
\end{aligned}
\label{eq:training_loss}
\end{equation}
The codec is frozen during denoiser training. Histories, generation regions, and future observations use explicit validity masks; outputs outside the generation region are excluded from the supervised reduction. The common budget used for architecture comparisons is specified in Sec.~\ref{sec:experiments}.

\providecommand{\motionmetrics}[5]{\begin{tabular}[c]{@{}c@{}}#1\,/\,#2\\#3\,/\,#4\\#5\end{tabular}}

\section{Experiments}
\label{sec:experiments}
In this section, we evaluate Motion Mamba 2.0 from three perspectives: generation quality, architectural effectiveness, and inference efficiency. We first compare short- and long-horizon generation on retargeted HiPHI, followed by short-horizon evaluations on BONES Seed and retargeted HumanML3D in Sec.~\ref{sec:main-results}. We then isolate the contributions of recurrent modeling and persistent content memory through architectural ablations in Sec.~\ref{sec:architecture-ablation}. Finally, we evaluate whether the proposed execution optimizations reduce inference latency without altering generated outputs or committed historical states in Sec.~\ref{sec:inference-efficiency}.

\begin{table*}[!tb]
\centering\captionsetup{width=\linewidth}
\caption{\textbf{Short- and Long-Horizon Motion Generation on Retargeted HiPHI.} CE is reported in cm, Cmd.\ and R@3 in percentages, and latency in ms per accepted window at batch size 1. R@3 uses 32 generated motion candidates. See App.~C for details.}
\label{tab:hiphi-paired-v4}
\begingroup
\fontsize{7.5}{8.5}\selectfont
\setlength{\tabcolsep}{1.4pt}
\resizebox{\linewidth}{!}{%
\begin{tabular}{@{}l*{24}{r}@{}}
\toprule
Method & \multicolumn{6}{c}{52 frames / 1 window} & \multicolumn{6}{c}{8.32 s / 4 windows} & \multicolumn{6}{c}{18.72 s / 9 windows} & \multicolumn{6}{c}{39.52 s / 19 windows} \\
\cmidrule(lr){2-7}\cmidrule(lr){8-13}\cmidrule(lr){14-19}\cmidrule(lr){20-25}
 & Root CE$\downarrow$ & End CE$\downarrow$ & Cmd.$\uparrow$ & R@3$\uparrow$ & L2-FID$\downarrow$ & Lat.$\downarrow$ & Root CE$\downarrow$ & End CE$\downarrow$ & Cmd.$\uparrow$ & R@3$\uparrow$ & L2-FID$\downarrow$ & Lat.$\downarrow$ & Root CE$\downarrow$ & End CE$\downarrow$ & Cmd.$\uparrow$ & R@3$\uparrow$ & L2-FID$\downarrow$ & Lat.$\downarrow$ & Root CE$\downarrow$ & End CE$\downarrow$ & Cmd.$\uparrow$ & R@3$\uparrow$ & L2-FID$\downarrow$ & Lat.$\downarrow$ \\
\midrule
DiP~\cite{DiP} & 57.07 & 60.45 & 40.78 & 25.66 & 0.402 & 73.6 & 62.58 & 66.16 & 31.59 & 11.08 & 0.444 & 68.7 & 77.69 & 80.88 & 20.50 & 12.84 & 0.487 & 68.7 & 96.99 & 99.26 & 18.45 & 12.45 & 0.490 & 66.8 \\
Go-to-Zero~\cite{GoToZero}
& 29.14 & 29.53 & 71.79 & 26.27 & 0.480 & 12344.9
& 39.65 & 40.19 & 40.63 & 18.17 & 0.514 & 12564.3
& 56.36 & 59.76 & 32.89 & 13.09 & 0.631 & 12403.2
& 86.47 & 88.91 & 25.03 & 12.11 & 0.743 & 12478.5 \\
DreamControl~\cite{DreamControl}
& 31.43 & 35.85 & 75.39 & 22.62 & 0.457 & 94.2
& 40.46 & 43.99 & 38.80 & 16.22 & 0.402 & 90.8
& 56.89 & 60.84 & 26.66 & 13.97 & 0.472 & 90.5
& 61.92 & 67.43 & 20.00 & 14.40 & 0.532 & 89.5 \\
MoLingo~\cite{MoLingo}
& 99.24 & 102.67 & 47.77 & 8.86 & 1.241 & 868.3
& 137.47 & 140.71 & 34.79 & 7.46 & 1.299 & 847.7
& 205.38 & 207.00 & 41.14 & 8.72 & 1.261 & 840.4
& 270.71 & 273.42 & 41.34 & 8.81 & 1.260 & 908.7 \\
HY-Motion~\cite{HYMotion}
& 30.76 & 33.54 & 55.70 & 16.67 & 0.400 & 324.1
& 43.94 & 45.97 & 37.37 & 12.47 & 0.492 & 274.0
& 65.06 & 66.75 & 31.96 & 14.52 & 0.629 & 290.5
& 95.53 & 96.63 & 27.61 & 13.47 & 0.760 & 346.8 \\
OMG~\cite{OMG}
& 58.18 & 60.59 & 46.76 & 25.93 & 0.448 & 173.9
& 82.83 & 85.97 & 36.41 & 18.84 & 0.428 & 170.2
& 115.50 & 118.11 & 29.85 & 20.96 & 0.443 & 168.3
& 161.76 & 163.49 & 24.52 & 19.61 & 0.473 & 167.8 \\
TACD~\cite{TACD}
& 21.81 & 25.55 & 77.27 & 28.44 & 0.428 & 203.0
& 30.49 & 34.66 & 43.71 & 18.26 & 0.422 & 224.9
& 40.19 & 44.16 & 34.88 & 20.63 & 0.418 & 191.6
& 54.43 & 59.40 & 28.03 & 17.81 & 0.496 & 208.0 \\
\midrule
PCMDM~\cite{PCMDM}
& 63.86 & 66.65 & 28.71 & 17.59 & 0.434 & 58.4
& 61.17 & 63.96 & 26.73 & 10.43 & 0.457 & 51.6
& 62.41 & 64.93 & 16.12 & 10.28 & 0.512 & 53.0
& 65.18 & 66.99 & 14.27 & 10.96 & 0.544 & 51.4 \\
FlowMDM~\cite{FlowMDM} & 56.67 & 58.01 & 46.25 & 14.29 & 0.400 & 109.9 & 72.80 & 73.93 & 35.00 & 10.61 & 0.437 & 105.2 & 112.44 & 113.81 & 26.35 & 10.57 & 0.443 & 107.2 & 154.71 & 156.35 & 24.12 & 11.07 & 0.454 & 105.7 \\
DartControl~\cite{DartControl} & 44.84 & 48.37 & 71.88 & 13.49 & 0.387 & \textbf{34.7} & 64.66 & 68.24 & 34.24 & 16.52 & 0.471 & \textbf{30.3} & 91.05 & 95.51 & 41.17 & 16.35 & 0.493 & \textbf{29.1} & 291.58 & 295.86 & 40.32 & 15.53 & 0.511 & \textbf{28.2} \\
Motion-Agent~\cite{Motion-Agent}
& 20.72 & 24.09 & 22.22 & 23.17 & 0.437 & 13806.4
& 31.76 & 33.25 & 26.69 & 12.36 & 0.496 & 13694.5
& 43.15 & 46.34 & 14.97 & 12.14 & 0.542 & 13319.4
& 61.36 & 62.49 & 15.15 & 12.02 & 0.745 & 13511.1 \\
MG-MotionLLM~\cite{MG-MotionLLM}
& 63.77 & 67.99 & 22.39 & 23.02 & 0.434 & 5632.7
& 98.31 & 99.67 & 25.48 & 9.22 & 0.455 & 5783.3
& 204.32 & 203.35 & 12.14 & 10.60 & 0.451 & 5861.8
& 340.84 & 341.01 & 11.52 & 11.60 & 0.431 & 5772.3 \\
IRG-MotionLLM~\cite{IRG-MotionLLM}
& 43.58 & 46.97 & 24.58 & 21.74 & 0.447 & 13329.8
& 70.64 & 72.89 & 23.86 & 11.93 & 0.472 & 13295.7
& 92.71 & 95.16 & 13.53 & 15.95 & 0.535 & 13198.4
& 113.23 & 131.52 & 13.04 & 13.15 & 0.751 & 12912.6 \\
MotionStreamer~\cite{MotionStreamer}
& 17.89 & 23.56 & 80.32 & 22.62 & 0.388 & 99.0 & 16.93 & 25.31 & 46.00 & 17.54 & 0.444 & 109.8 & 6.17 & 18.07 & 26.87 & 15.83 & 0.542 & 119.2 & 7.78 & 21.67 & 18.85 & 13.02 & 0.645 & 159.8 \\
CMDM~\cite{CMDM} & 24.65 & 27.16 & 84.02 & 23.41 & \textbf{0.337} & 157.8 & 38.93 & 42.65 & 34.30 & 14.65 & 0.454 & 186.7 & 49.35 & 53.04 & 22.80 & 14.49 & 0.549 & 230.4 & 54.68 & 59.68 & 15.36 & 12.10 & 0.657 & 311.9 \\
ActionPlan~\cite{ActionPlan}
& 69.72 & 83.69 & 52.79 & 9.13 & 0.862 & 72.4
& 93.89 & 106.81 & 36.76 & 4.70 & 1.047 & 68.8
& 281.64 & 279.40 & 40.38 & 6.43 & 1.067 & 85.3
& 799.60 & 790.99 & 40.21 & 6.16 & 1.120 & 110.1 \\ 
FloodDiffusion 2~\cite{FD2} & 39.12 & 41.54 & 63.51 & 20.63 & 0.403 & 196.8 & 67.72 & 70.42 & 36.03 & 12.44 & 0.433 & 190.8 & 90.70 & 93.48 & 26.95 & 13.55 & 0.465 & 186.9 & 115.02 & 117.86 & 22.34 & 12.14 & 0.497 & 183.7 \\
ARDY~\cite{ARDY} & 5.89 & 9.70 & 86.77 & 23.54 & 0.425 & 95.6 & 7.44 & 10.27 & 45.83 & 13.76 & 0.438 & 90.1 & 8.15 & 11.06 & 30.89 & 17.14 & 0.467 & 91.5 & 10.09 & 14.84 & 23.53 & 15.32 & 0.489 & 105.1 \\
\midrule
\rowcolor{gray!10}
Motion Mamba~\cite{Motion_Mamba} & 50.58 & 53.39 & 54.75 & 19.84 & 0.406 & 309.2 & 56.81 & 59.83 & 44.06 & 17.98 & 0.414 & 317.3 & 78.67 & 82.37 & 31.46 & 17.29 & 0.420 & 315.1 & 125.74 & 130.17 & 25.77 & 17.18 & 0.482 & 310.5 \\
\rowcolor{cyan!20}
\textbf{Motion Mamba 2.0} & \textbf{2.78} & \textbf{7.19} & \textbf{86.96} & \textbf{35.58} & 0.410 & 283.7 & \textbf{2.53} & \textbf{7.11} & \textbf{48.82} & \textbf{21.35} & \textbf{0.401} & 267.5 & \textbf{2.87} & \textbf{8.45} & \textbf{47.11} & \textbf{23.05} & \textbf{0.402} & 284.7 & \textbf{7.55} & \textbf{13.64} & \textbf{46.05} & \textbf{23.68} & \textbf{0.409} & 333.4 \\
\bottomrule
\end{tabular}%
}
\endgroup
\end{table*}

\subsection{Experimental Setups}
\label{sec:experimental-setup}

\resultheading{Training Procedure}
Motion Mamba 2.0 is jointly trained with Mamba-3 and persistent content memory (PCM) enabled from initialization. The G1 motion codec remains frozen, while cached text features provide conditioning. Training minimizes the
four-term objective in  Eq.~\eqref{eq:training_loss}.
We use AdamW for 100,000 updates with a global batch size of 512 throughout training. The learning rate increases linearly from $10^{-6}$ at update 1 to $5\times10^{-5}$ at update 500, followed by cosine decay to $10^{-5}$ at update 100,000. Gradients are clipped to a maximum norm of $1.0$. The generation length is fixed at 52 frames. History/future lengths $(H,F)$ are sampled from $(0,0)$, $(52,0)$, $(196,0)$, $(0,196)$, $(52,144)$, and $(144,52)$, with respective frequencies of $2,5,4,2,4,3$ per 20 updates. The maximum sampled training history is therefore 196 frames (7.84\,s).

Training samples are drawn from BONES Seed, retargeted HumanML3D, and retargeted HiPHI in a 70:15:15 ratio. Each sample contains a 52-frame generation window with historical motion conditions constructed from ground-truth sequences and sampled future constraints; training does not autoregressively roll out generated histories. Text and spatial condition dropout are both set to $0.1$, with an empty spatial-condition probability of $0.2$. For fair comparison, all baselines and architectural variants are independently trained on the same G1 corpus
under the matched 100,000-update budget and batch-size schedule. All evaluations use the resulting G1-adapted checkpoints rather than the original pretrained releases. Further details are provided in App.~C.

\resultheading{Evaluation Protocol}
Motion is generated at 25\,Hz in 52-frame windows
(2.08\,s). Retargeted HiPHI~\cite{HiPHI} is evaluated
over 1, 4, 9, and 19 windows (2.08--39.52\,s), while
BONES Seed~\cite{BONES} and retargeted
HumanML3D~\cite{HumanML3D} use 1 and 4 windows.
For each horizon and method, 128 held-out initialization
families and three noise seeds yield 384 rollouts under
paired evaluation conditions. HiPHI also shares initial
histories, spatial targets, and noise seeds across
horizons, while instruction schedules and terminal
deadlines differ. Each trial specifies a root target and
one wrist or ankle target, available from rollout onset
with a terminal deadline; only the active text instruction
is provided during generation. Sampling uses 10-step
deterministic DDIM ($\eta=0$)~\cite{DDIM} with uniformly
spaced timesteps and text/spatial guidance scales of
$(2,2)$. Further details are provided in
App.~C.

\resultheading{Evaluation Metrics}
Root CE measures horizontal root-position error in
the world $X$--$Z$ plane, while End CE measures
the 3D positional error of the constrained wrist
or ankle at the terminal deadline, both in centimeters.
Command matching (Cmd.) evaluates coarse root-motion
state agreement. Text--motion alignment is evaluated
by R@3 using independently trained, frozen evaluators
for each benchmark. HiPHI uses 32 generated candidates,
whereas BONES Seed and HumanML3D use one matched
generated candidate and 31 fixed real-motion distractors.
R@3 is therefore comparable within, but not across 
benchmarks. L2-FID compares distributions of
L2-normalized 256-dimensional motion features from
phase-paired 52-frame windows, including those in
extended rollouts. Latency is measured per accepted
window at batch size one. Metric aggregation and
reference protocols are detailed in
App.~C.

\subsection{Main Results}
\label{sec:main-results}

\resultheading{Results on HiPHI}
Tab.~\ref{tab:hiphi-paired-v4} compares four horizons from 2.08 to 39.52\,s. Motion Mamba 2.0 achieves the lowest Root CE at every duration and the lowest or tied-lowest End CE. At 18.72\,s, Root/End CE reach 2.87/8.45\,cm, compared with 6.17/18.07\,cm for MotionStreamer and 8.15/11.06\,cm for ARDY. At 39.52\,s, our method achieves 7.55/13.64\,cm, reducing Root CE by 3.0\% relative to MotionStreamer (7.78\,cm) and 
outperforming the best competing End CE of 14.84\,cm from ARDY by 8.1\%.
These results demonstrate accurate terminal constraint satisfaction over extended rollouts.
A qualitative comparison with ARDY under a distant spatial constraint is provided in App.~C.6.

Motion Mamba 2.0 also achieves the highest Cmd.\ and R@3 across all four horizons, supporting command agreement and text--motion retrieval under the evaluated protocol. At 39.52\,s, Cmd.\ and R@3 reach 46.05\% and 23.68\%, compared with 41.34\% for MoLingo and 19.61\% for OMG, respectively. The method obtains the lowest L2-FID at 8.32, 18.72, and 39.52\,s (0.401, 0.402, and 0.409); CMDM leads at 2.08\,s (0.337 versus 0.410), and the 8.32\,s advantage over DreamControl (0.402) is small. Together, these results demonstrate strong terminal control alongside long-horizon command agreement and retrieval performance.

\begin{table*}[!tb]
\centering\captionsetup{width=0.9\linewidth}
\caption{\textbf{Short-Horizon Motion Generation on BONES Seed.} CE is reported in cm, Cmd.\ and R@3 in percentages, and latency in ms per accepted window at batch size 1. R@3 uses one matched generated motion and 31 fixed real-motion distractors. Full protocol: App.~C.}
\label{tab:bones-seed}
\begingroup
\fontsize{7.5}{8.5}\selectfont
\setlength{\tabcolsep}{2.5pt}
\resizebox{0.9\linewidth}{!}{%
\begin{tabular}{@{}l*{12}{r}@{}}
\toprule
Method & \multicolumn{6}{c}{52 frames (1 window)} & \multicolumn{6}{c}{8.32 s (4 windows)} \\
\cmidrule(lr){2-7}\cmidrule(lr){8-13}
 & Root CE$\downarrow$ & End CE$\downarrow$ & Cmd.$\uparrow$ & $R@3\uparrow$ & L2-FID$\downarrow$ & Lat.$\downarrow$ & Root CE$\downarrow$ & End CE$\downarrow$ & Cmd.$\uparrow$ & $R@3\uparrow$ & L2-FID$\downarrow$ & Lat.$\downarrow$ \\
\midrule
DiP~\cite{DiP} & 60.28 & 63.68 & 44.61 & 18.84 & 0.461 & 74.7 & 61.21 & 63.68 & 36.02 & 32.30 & 0.640 & 70.3 \\
Go-to-Zero~\cite{GoToZero}
& 30.80 & 32.50 & 72.39 & 17.55 & 0.487 & 12275.5
& 43.19 & 41.27 & 39.46 & 12.23 & 0.613 & 12316.8 \\
DreamControl~\cite{DreamControl} & 22.70 & 27.33 & 96.64 & 40.60 & 0.608 & 93.1 & 36.06 & 40.20 & 54.93 & 26.49 & 0.522 & 92.1 \\
MoLingo~\cite{MoLingo} & 93.85 & 95.66 & 97.50 & 32.64 & 1.283 & 846.7 & 122.43 & 124.64 & 47.82 & 15.15 & 1.207 & 841.7 \\
HY-Motion~\cite{HYMotion}
& 38.94 & 31.32 & 59.93 & 18.51 & 0.456 & 282.3
& 51.94 & 54.12 & 32.85 & 11.80 & 0.583 & 276.9 \\
OMG~\cite{OMG} & 76.58 & 77.97 & 46.18 & 26.39 & \textbf{0.345} & 173.4 & 123.93 & 125.52 & 31.48 & 30.24 & 0.544 & 173.9 \\
TACD~\cite{TACD}
& 32.70 & 35.43 & 68.36 & 35.72 & 0.483 & 201.7
& 35.86 & 39.61 & 40.50 & 28.80 & 0.555 & 192.3 \\
\midrule
PCMDM~\cite{PCMDM} & 65.98 & 67.49 & 41.00 & 14.97 & 0.437 & 58.8 & 66.77 & 68.26 & 29.77 & 24.85 & 0.600 & 53.5 \\
FlowMDM~\cite{FlowMDM} & 76.32 & 77.72 & 38.10 & 16.19 & 0.382 & 113.6 & 92.96 & 94.37 & 30.62 & 24.56 & 0.518 & 107.3 \\
DartControl~\cite{DartControl} & 46.51 & 46.25 & 96.60 & 48.66 & 0.602 & \textbf{35.2} & 272.86 & 272.43 & \textbf{61.79} & 27.90 & 0.668 & \textbf{30.8} \\
Motion-Agent~\cite{Motion-Agent} & 33.79 & 35.23 & 32.37 & 36.90 & 0.538 & 13461.2 & 48.05 & 50.89 & 26.53 & 25.98 & 0.517 & 13624.0 \\
MG-MotionLLM~\cite{MG-MotionLLM} & 82.91 & 86.77 & 54.57 & 37.50 & 0.460 & 5623.9 & 328.11 & 329.51 & 25.83 & 30.01 & 0.469 & 5745.0 \\
IRG-MotionLLM~\cite{IRG-MotionLLM} & 49.68 & 42.81 & 36.62 & 38.61 & 0.501 & 14497.4 & 69.05 & 76.78 & 29.16 & 30.38 & 0.621 & 13868.3 \\
MotionStreamer~\cite{MotionStreamer} & 15.08 & 22.29 & 88.32 & 53.99 & 0.570 & 102.7 & 12.97 & 19.54 & 43.01 & 26.63 & 0.542 & 106.2 \\
CMDM~\cite{CMDM} & 32.66 & 34.71 & 85.90 & 48.65 & 0.439 & 182.1 & 48.56 & 52.06 & 37.15 & 25.07 & 0.579 & 197.3 \\
ActionPlan~\cite{ActionPlan} & 77.76 & 89.85 & \textbf{100.00} & 44.08 & 1.237 & 71.8 & 223.18 & 223.45 & 54.59 & 12.76 & 1.405 & 68.5 \\
FloodDiffusion 2~\cite{FD2} & 52.50 & 54.09 & 62.26 & 34.98 & 0.409 & 199.4 & 71.05 & 73.28 & 35.62 & 26.64 & 0.565 & 191.3 \\
ARDY~\cite{ARDY} & 5.57 & 9.63 & 83.14 & 48.74 & 0.596 & 98.1 & 4.96 & 10.16 & 45.10 & 34.26 & 0.568 & 102.7 \\
\midrule
\rowcolor{gray!10}
Motion Mamba~\cite{Motion_Mamba} & 43.29 & 44.65 & 77.35 & 56.47 & 0.560 & 315.4 & 58.64 & 60.83 & 42.31 & 29.15 & 0.591 & 301.8 \\
\rowcolor{cyan!20}
\textbf{Motion Mamba 2.0} & \textbf{2.10} & \textbf{7.09} & 86.92 & \textbf{59.38} & 0.573 & 290.1 & \textbf{1.86} & \textbf{6.86} & 51.42 & \textbf{35.58} & \textbf{0.459} & 271.1 \\
\bottomrule
\end{tabular}%
}
\endgroup
\end{table*}

\begin{table*}[!tb]
\centering\captionsetup{width=0.9\linewidth}
\caption{\textbf{Short-Horizon Motion Generation on Retargeted HumanML3D.} CE is reported in cm, Cmd.\ and R@3 in percentages, and latency in ms per accepted window at batch size 1. R@3 uses one matched generated motion and 31 fixed real-motion distractors. Full protocol: App.~C.}
\label{tab:humanml3d}
\begingroup
\fontsize{7.5}{8.5}\selectfont
\setlength{\tabcolsep}{2.5pt}
\resizebox{0.9\linewidth}{!}{%
\begin{tabular}{@{}l*{12}{r}@{}}
\toprule
Method & \multicolumn{6}{c}{52 frames (1 window)} & \multicolumn{6}{c}{8.32 s (4 windows)} \\
\cmidrule(lr){2-7}\cmidrule(lr){8-13}
 & Root CE$\downarrow$ & End CE$\downarrow$ & Cmd.$\uparrow$ & $R@3\uparrow$ & L2-FID$\downarrow$ & Lat.$\downarrow$ & Root CE$\downarrow$ & End CE$\downarrow$ & Cmd.$\uparrow$ & $R@3\uparrow$ & L2-FID$\downarrow$ & Lat.$\downarrow$ \\
\midrule
DiP~\cite{DiP} & 49.72 & 54.03 & 64.59 & 84.46 & 0.623 & 76.1 & 52.78 & 55.17 & 41.04 & 47.95 & 0.706 & 69.4 \\
Go-to-Zero~\cite{GoToZero}
& 29.11 & 21.71 & 52.30 & 66.90 & 0.462 & 12272.0
& 34.78 & 31.96 & 34.15 & 40.31 & 0.541 & 12351.6 \\
DreamControl~\cite{DreamControl} & 22.06 & 27.14 & 97.24 & 87.98 & 1.100 & 98.1 & 36.02 & 39.27 & 57.48 & 45.76 & 0.737 & 94.1 \\
MoLingo~\cite{MoLingo} & 93.94 & 99.88 & 97.82 & 79.90 & 1.551 & 881.4 & 130.90 & 135.04 & 47.82 & 43.55 & 1.514 & 862.7 \\
HY-Motion~\cite{HYMotion}
& 35.46 & 37.00 & 51.78 & 61.83 & 0.383 & 317.5
& 47.05 & 48.94 & 33.24 & 31.78 & 0.579 & 292.0 \\
OMG~\cite{OMG}
& 75.87 & 76.51 & 72.36 & 72.70 & \textbf{0.319} & 175.6
& 136.38 & 137.49 & 36.37 & 32.52 & 0.564 & 170.8 \\
TACD~\cite{TACD}
& 25.75 & 27.84 & 78.68 & 11.98 & 0.329 & 204.3
& 34.66 & 37.32 & 42.67 & 17.42 & 0.490 & 195.6 \\
\midrule
PCMDM~\cite{PCMDM} & 51.98 & 54.63 & 64.60 & 74.74 & 0.592 & 60.1 & 51.09 & 53.63 & 35.84 & 40.38 & 0.729 & 54.8 \\
FlowMDM~\cite{FlowMDM} & 67.89 & 68.95 & 65.54 & 70.40 & 0.364 & 109.8 & 104.39 & 105.96 & 38.20 & 40.90 & 0.554 & 104.0 \\
DartControl~\cite{DartControl} & 67.17 & 64.09 & 96.96 & 87.50 & 0.528 & \textbf{39.3} & 238.78 & 235.93 & \textbf{61.04} & 48.35 & 0.732 & \textbf{29.7} \\
Motion-Agent~\cite{Motion-Agent}
& 22.20 & 25.39 & 52.55 & 65.49 & 0.574 & 13661.4 & 34.60 & 35.43 & 26.81 & 44.63 & 0.611 & 13264.4 \\
MG-MotionLLM~\cite{MG-MotionLLM} & 60.45 & 64.31 & 53.16 & 67.06 & 0.558 & 5590.8 & 160.42 & 159.49 & 25.71 & 40.33 & 0.554 & 5743.3 \\
IRG-MotionLLM~\cite{IRG-MotionLLM}
& 45.69 & 43.30 & 51.31 & 63.82 & 0.490 & 13556.0 & 65.03 & 67.18 & 24.09 & 42.15 & 0.618 & 13468.2 \\
MotionStreamer~\cite{MotionStreamer}
& 15.24 & 20.23 & 91.32 & \textbf{96.40} & 0.463 & 106.7 & 12.16 & 18.28 & 47.63 & 44.39 & 0.557 & 105.4 \\
CMDM~\cite{CMDM} & 32.61 & 34.20 & 92.80 & 89.76 & 0.375 & 129.0 & 47.30 & 49.82 & 44.37 & 46.08 & 0.572 & 129.9 \\
ActionPlan~\cite{ActionPlan} & 76.26 & 90.77 & \textbf{100.00} & 40.13 & 1.737 & 74.8 & 221.90 & 222.02 & 55.11 & 40.27 & 2.005 & 70.2 \\
FloodDiffusion 2~\cite{FD2} & 52.15 & 53.78 & 82.43 & 77.86 & 0.334 & 164.1 & 71.64 & 74.09 & 42.51 & 42.08 & 0.593 & 153.9 \\
ARDY~\cite{ARDY} & 5.22 & 8.85 & 87.55 & 86.85 & 0.430 & 100.6 & 4.75 & 9.82 & 52.28 & 48.21 & 0.491 & 91.1 \\
\midrule
\rowcolor{gray!10}
Motion Mamba~\cite{Motion_Mamba} & 38.42 & 40.17 & 85.74 & 84.57 & 0.747 & 309.8 & 59.26 & 61.31 & 54.47 & 40.14 & 0.653 & 313.9 \\
\rowcolor{cyan!20}
\textbf{Motion Mamba 2.0} & \textbf{2.18} & \textbf{6.73} & 88.07 & 84.29 & 0.459 & 286.0 & \textbf{2.17} & \textbf{7.03} & 58.81 & \textbf{49.72} & \textbf{0.490} & 271.2 \\
\bottomrule
\end{tabular}%
}
\endgroup
\end{table*}

\resultheading{Results on BONES Seed}
Tab.~\ref{tab:bones-seed} evaluates short-horizon generation on native G1 trajectories from BONES Seed. Motion Mamba 2.0 achieves the lowest Root/End CE and highest R@3 at both durations. At 8.32\,s, Root CE decreases to 1.86\,cm from 4.96\,cm for ARDY, a relative reduction of 62.5\%, while End CE decreases from 10.16 to 6.86\,cm. The corresponding Cmd.\ and R@3 are 51.42\% and 35.58\%, respectively; command matching is not the highest among compared methods.

For distributional similarity, our method obtains the lowest L2-FID at 8.32\,s (0.459), whereas OMG leads at 2.08\,s (0.345 versus 0.573). DartControl remains the fastest method. These results confirm the spatial-control and R@3 advantages on native G1 motions, while coarse command matching and single-window distributional quality are not uniformly best.

\resultheading{Results on HumanML3D}
Tab.~\ref{tab:humanml3d} reports results on retargeted HumanML3D at 2.08 and 8.32\,s, within the available source duration. Motion Mamba 2.0 achieves the lowest Root/End CE at both horizons. At 8.32\,s, Root/End CE reach 2.17/7.03\,cm, compared with 4.75/9.82\,cm for ARDY, a 54.3\% reduction in Root CE. It also obtains the highest four-window R@3 of 49.72\%, exceeding DartControl (48.35\%). At 2.08\,s, its R@3 of 84.29\% remains below MotionStreamer (96.40\%), indicating a difference between precise constraint control and single-window text retrieval.

OMG obtains the lowest L2-FID at 2.08\,s (0.319), whereas our method narrowly leads at 8.32\,s (0.490 versus 0.491 for ARDY). Together with BONES Seed, these results support accurate spatial control on both retargeted human motions and native G1 references without requiring extended rollout durations, while showing that semantic metrics vary across benchmarks.

\subsection{Architectural Ablations}
\label{sec:architecture-ablation}

We assess recurrent modeling and PCM through five configurations: \emph{Minimal} (without the Mamba branch), \emph{Native} (Mamba without PCM), \emph{Native + PCM (w/o text)}, \emph{Native + PCM (w/o delta)}, and the complete \emph{Motion Mamba 2.0}. The last two partial-PCM variants ablate text conditioning and the explicit state-difference feature in the PCM writing path, respectively. The reported variants use the matched 100,000-update training and evaluation protocol. Tabs.~\ref{tab:hiphi-ablation-v4}, \ref{tab:bones-seed-ablation}, and \ref{tab:humanml3d-ablation} report the results on HiPHI, BONES Seed, and HumanML3D.

\begin{table*}[!tb]
\centering\captionsetup{width=\linewidth}
\caption{\textbf{Architectural Ablation on Retargeted HiPHI.} Training budgets and execution paths are specified in App.~C. Metrics and units follow Tab.~\ref{tab:hiphi-paired-v4}.}
\label{tab:hiphi-ablation-v4}
\begingroup
\fontsize{7.5}{8.5}\selectfont
\setlength{\tabcolsep}{1.4pt}
\resizebox{\linewidth}{!}{%
\begin{tabular}{@{}l*{24}{r}@{}}
\toprule
Method & \multicolumn{6}{c}{52 frames / 1 window} & \multicolumn{6}{c}{8.32 s / 4 windows} & \multicolumn{6}{c}{18.72 s / 9 windows} & \multicolumn{6}{c}{39.52 s / 19 windows} \\
\cmidrule(lr){2-7}\cmidrule(lr){8-13}\cmidrule(lr){14-19}\cmidrule(lr){20-25}
 & Root CE$\downarrow$ & End CE$\downarrow$ & Cmd.$\uparrow$ & R@3$\uparrow$ & L2-FID$\downarrow$ & Lat.$\downarrow$ & Root CE$\downarrow$ & End CE$\downarrow$ & Cmd.$\uparrow$ & R@3$\uparrow$ & L2-FID$\downarrow$ & Lat.$\downarrow$ & Root CE$\downarrow$ & End CE$\downarrow$ & Cmd.$\uparrow$ & R@3$\uparrow$ & L2-FID$\downarrow$ & Lat.$\downarrow$ & Root CE$\downarrow$ & End CE$\downarrow$ & Cmd.$\uparrow$ & R@3$\uparrow$ & L2-FID$\downarrow$ & Lat.$\downarrow$ \\
\midrule
Minimal & 3.86 & 12.44 & \textbf{89.69} & 34.62 & 0.371 & 509.0 & 5.92 & 13.52 & \textbf{52.27} & 20.14 & 0.414 & 536.5 & 18.70 & 23.58 & 38.27 & 18.07 & 0.427 & 557.6 & 60.00 & 64.86 & 35.27 & 17.26 & 0.439 & 552.9 \\
Native & 4.49 & 10.88 & 88.14 & 32.67 & \textbf{0.368} & \textbf{233.4} & 4.60 & 9.84 & 51.69 & 19.64 & 0.462 & \textbf{200.6} & 6.21 & 11.44 & 38.21 & 19.08 & 0.441 & \textbf{212.5} & 9.36 & 15.54 & 32.20 & 17.44 & 0.416 & \textbf{258.7} \\
Native + PCM (w/o text) & 3.19 & 7.65 & 87.01 & 32.01 & 0.423 & 286.5 & 3.05 & 7.81 & 47.60 & 19.41 & 0.419 & 267.8 & 2.99 & 9.02 & 38.10 & 21.06 & 0.437 & 292.8 & 8.62 & 14.62 & 33.34 & 21.18 & 0.467 & 336.5 \\
Native + PCM (w/o delta) & 3.06 & 7.41 & 86.36 & 32.67 & 0.433 & 287.6 & 2.97 & 7.19 & 47.39 & 19.76 & 0.421 & 263.1 & 3.08 & 8.99 & 37.18 & 20.35 & 0.459 & 284.7 & 8.30 & 14.18 & 32.65 & 21.11 & 0.488 & 329.8 \\
Motion Mamba 2.0 & \textbf{2.78} & \textbf{7.19} & 86.96 & \textbf{35.58} & 0.410 & 283.7 & \textbf{2.53} & \textbf{7.11} & 48.82 & \textbf{21.35} & \textbf{0.401} & 267.5 & \textbf{2.87} & \textbf{8.45} & \textbf{47.11} & \textbf{23.05} & \textbf{0.402} & 284.7 & \textbf{7.55} & \textbf{13.64} & \textbf{46.05} & \textbf{23.68} & \textbf{0.409} & 333.4 \\
\bottomrule
\end{tabular}%
}
\endgroup
\end{table*}

\begin{table*}[!tb]
\centering\captionsetup{width=0.9\linewidth}
\caption{\textbf{Architectural ablation on BONES Seed.} Training budgets and execution paths are specified in App.~C. Units and $R@3$ follow Tab.~\ref{tab:bones-seed}.}
\label{tab:bones-seed-ablation}
\begingroup
\fontsize{7.5}{8.5}\selectfont
\setlength{\tabcolsep}{2.5pt}
\resizebox{0.9\linewidth}{!}{%
\begin{tabular}{@{}l*{12}{r}@{}}
\toprule
Method & \multicolumn{6}{c}{52 frames (1 window)} & \multicolumn{6}{c}{8.32 s (4 windows)} \\
\cmidrule(lr){2-7}\cmidrule(lr){8-13}
 & Root CE$\downarrow$ & End CE$\downarrow$ & Cmd.$\uparrow$ & $R@3\uparrow$ & L2-FID$\downarrow$ & Lat.$\downarrow$ & Root CE$\downarrow$ & End CE$\downarrow$ & Cmd.$\uparrow$ & $R@3 \uparrow$ & L2-FID$\downarrow$ & Lat.$\downarrow$ \\
\midrule
Minimal & 6.76 & 12.82 & \textbf{88.09} & 57.21 & 0.561 & 490.8 & 11.10 & 17.17 & 48.02 & 31.66 & 0.656 & 539.1 \\
Native & 5.17 & 10.75 & 87.16 & 56.93 & \textbf{0.552} & \textbf{219.8} & 5.46 & 11.29 & \textbf{52.06} & 35.15 & 0.516 & \textbf{210.5} \\
Native + PCM (w/o text) & 3.34 & 8.48 & 87.25 & 54.82 & 0.568 & 286.2 & 3.53 & 8.21 & 44.16 & 30.71 & 0.473 & 270.4 \\
Native + PCM (w/o delta) & 3.22 & 8.61 & 86.58 & 54.56 & 0.575 & 290.2 & 3.61 & 8.74 & 43.75 & 30.47 & 0.481 & 268.6 \\
Motion Mamba 2.0 & \textbf{2.10} & \textbf{7.09} & 86.92 & \textbf{59.38} & 0.573 & 290.1 & \textbf{1.86} & \textbf{6.86} & 51.42 & \textbf{35.58} & \textbf{0.459} & 271.1 \\
\bottomrule
\end{tabular}%
}
\endgroup
\end{table*}

\begin{table*}[!tb]
\centering\captionsetup{width=0.9\linewidth}
\caption{\textbf{Architectural ablation on Retargeted HumanML3D.} Training budgets and execution paths are specified in App.~C. Units and $R@3$ follow Tab.~\ref{tab:humanml3d}.}
\label{tab:humanml3d-ablation}
\begingroup
\fontsize{7.5}{8.5}\selectfont
\setlength{\tabcolsep}{2.5pt}
\resizebox{0.9\linewidth}{!}{%
\begin{tabular}{@{}l*{12}{r}@{}}
\toprule
Method & \multicolumn{6}{c}{52 frames (1 window)} & \multicolumn{6}{c}{8.32 s (4 windows)} \\
\cmidrule(lr){2-7}\cmidrule(lr){8-13}
 & Root CE$\downarrow$ & End CE$\downarrow$ & Cmd.$\uparrow$ & $R@3\uparrow$ & L2-FID$\downarrow$ & Lat.$\downarrow$ & Root CE$\downarrow$ & End CE$\downarrow$ & Cmd.$\uparrow$ & $R@3 \uparrow$ & L2-FID$\downarrow$ & Lat.$\downarrow$ \\
\midrule
Minimal & 8.31 & 13.05 & \textbf{90.56} & \textbf{92.97} & \textbf{0.410} & 512.0 & 14.31 & 17.49 & 55.56 & \textbf{53.42} & 0.583 & 544.8 \\
Native & 6.58 & 11.48 & 89.37 & 88.72 & 0.448 & \textbf{230.4} & 6.25 & 11.07 & 55.86 & 53.27 & 0.579 & \textbf{219.1} \\
Native + PCM (w/o text) & 3.29 & 8.37 & 88.54 & 85.24 & 0.444 & 287.9 & 3.40 & 8.67 & 51.04 & 43.07 & \textbf{0.462} & 273.1 \\
Native + PCM (w/o delta) & 3.87 & 8.36 & 88.63 & 83.59 & 0.442 & 288.8 & 3.82 & 8.54 & 51.18 & 42.43 & 0.477 & 267.1 \\
Motion Mamba 2.0 & \textbf{2.18} & \textbf{6.73} & 88.07 & 84.29 & 0.459 & 286.0 & \textbf{2.17} & \textbf{7.03} & \textbf{58.81} & 49.72 & 0.490 & 271.2 \\
\bottomrule
\end{tabular}%
}
\endgroup
\end{table*}

\resultheading{Effect of recurrent modeling}
The benefit of recurrent modeling becomes more pronounced as the generation horizon increases (Tab.~\ref{tab:hiphi-ablation-v4}). At 39.52\,s, introducing Mamba reduces Root CE from 60.00 to 9.36\,cm and End CE from 64.86 to 15.54\,cm compared with Minimal; corresponding single-window differences are much smaller. Similar improvements in constraint errors are observed on BONES Seed and HumanML3D at 8.32\,s. These results support the contribution of recurrent state propagation to extended spatial control. The gain is not uniform across metrics, as Native has lower Cmd.\ than Minimal at 39.52\,s (32.20\% versus 35.27\%).

\resultheading{Contribution of persistent content memory}
Adding PCM to Native consistently improves Root CE and End CE across the three benchmarks (Tabs.~\ref{tab:hiphi-ablation-v4}--\ref{tab:humanml3d-ablation}). On HiPHI at 39.52\,s, PCM reduces Root CE from 9.36 to 7.55\,cm (19.3\%) and End CE from 15.54 to 13.64\,cm (12.2\%), while increasing Cmd.\ from 32.20\% to 46.05\%. R@3 rises from 17.44\% to 23.68\%, and L2-FID decreases from 0.416 to 0.409. These gains establish its contribution to long-horizon control and retrieval.

At the same 39.52\,s setting, excluding text conditioning increases Root/End CE to 8.62/14.62\,cm and decreases Cmd./R@3 to 33.34\%/21.18\%; excluding the state-difference feature gives 8.30/14.18\,cm and 32.65\%/21.11\%, respectively. The complete writer also improves Cmd. relative to both partial-PCM variants at 18.72\,s (47.11\% versus 38.10\% and 37.18\%). Both partial variants retain better long-protocol Root/End CE than Native, supporting the utility of the underlying compressed-state-to-attention interface. On BONES Seed and retargeted HumanML3D at 8.32\,s, the complete model reduces Root CE from 5.46 to 1.86\,cm and from 6.25 to 2.17\,cm, respectively. Improvements are not uniform across metrics: some short-horizon Cmd., R@3, and L2-FID scores favor other variants. These results support complementary roles for state changes and text-conditioned writing in the evaluated long-horizon spatial-control and command-matching tasks, without proving exact retrieval of arbitrary evicted motion events.

\subsection{Inference Efficiency}
\unskip
\label{sec:inference-efficiency}
\label{sec:pcm-efficiency}

We evaluate whether the proposed execution optimizations reduce generation cost while preserving the complete PCM model. All execution variants use identical frozen weights. This study separates execution-flow simplification, local A/DT kernel fusion, and CUDA Graph replay; it does not change the denoising schedule or memory architecture.

\resultheading{Measurement protocol}
Experiments use one RTX 4090, batch size one, ten denoising steps, three classifier-free guidance branches with scales $(2,2)$, and four windows of history. Attention remains FP32, the existing BF16 boundaries inside Mamba are retained, and TF32 is disabled. Each timing includes the complete accepted 52-frame window, from state restoration through decoding and CPU snapshot/export. Text embeddings are preloaded; loading, warmup, and graph capture are excluded. We collect 30 matched window pairs in each of five independent processes. Paired variants start from identical session and random states. Confidence intervals use 10,000 process-block bootstrap resamples. The timing boundary and numerical checks are detailed in App.~B.4.

\resultheading{Complete-window performance}
As shown in Tab.~\ref{tab:pcm-ablation}, the complete scheme reduces p50 latency from 1685.35 to 276.14\,ms and p95 from 2471.65 to 345.82\,ms, reductions of 83.6\% and 86.0\%, respectively. The median paired speedup is $6.138\times$, with a 95\% confidence interval of $[6.094,6.290]$. For a 2.08-s motion window, the optimized p50 corresponds to a real-time factor of 0.133. This measures steady-state generation, rather than cold-start or command-to-visible-motion latency.

Execution-flow simplification reduces p50 to 1102.82\,ms by removing unused pending-state materialization and diagnostic synchronization. Local fusion reduces it further to 1069.52\,ms. Joint capture and replay of the three CFG branches account for the larger improvement by reducing repeated host dispatch, reaching 285.20\,ms without fusion; adding fusion reaches 276.14\,ms, with a further paired speedup of $1.029\times$. Thus, the complete speedup cannot be attributed to the fused kernel alone.

\begin{figure}[!t]
\centering
\begin{tikzpicture}
\begin{axis}[
  width=\linewidth,height=4.8cm,
  ybar,bar width=9pt,
  xmin=-.55,xmax=2.55,ymin=0,ymax=128,
  xtick={0,1,2},
  xticklabels={GPU kernels,Host launches,Synchronizations},
  ytick={0,25,50,75,100},
  ylabel={Relative count (\%)},
  axis lines=left,axis line style={line width=.45pt},
  tick style={line width=.4pt},
  ymajorgrids=true,grid style={black!10,line width=.35pt},
  font=\rmfamily\fontsize{8}{9.2}\selectfont,
  tick label style={font=\rmfamily\fontsize{7.5}{9}\selectfont},
  label style={font=\rmfamily\fontsize{8}{9.2}\selectfont},
  legend style={at={(.98,.98)},anchor=north east,draw=none,
    fill=none,legend columns=2,column sep=6pt,
    font=\rmfamily\fontsize{7.5}{9}\selectfont},
  enlarge x limits=false,clip=false]
\addplot[area legend,fill=pcmgray,draw=none] coordinates {(0,100) (1,100) (2,100)};
\addplot[area legend,fill=pcmblue,draw=none,
  nodes near coords={\pgfmathprintnumber[fixed,precision=1,zerofill]{\pgfplotspointmeta}},point meta=explicit,
  every node near coord/.append style={font=\rmfamily\fontsize{7.5}{9}\selectfont,
    text=black,anchor=south},
  nodes near coords align={vertical}]
  coordinates {(0,51.6724429757) [51.7] (1,6.5877269418) [6.6] (2,13.0468903276) [13.0]};
\legend{Eager,Optimized}
\end{axis}
\end{tikzpicture}
\caption{\textbf{PCM execution profile before and after optimization, normalized to eager execution.} Ordinary host launches exclude CUDA Graph replay calls. Counts come from separate profiling runs and are independent of the latency measurements in Tab.~\ref{tab:pcm-ablation}.}
\label{fig:pcm-profile}
\end{figure}
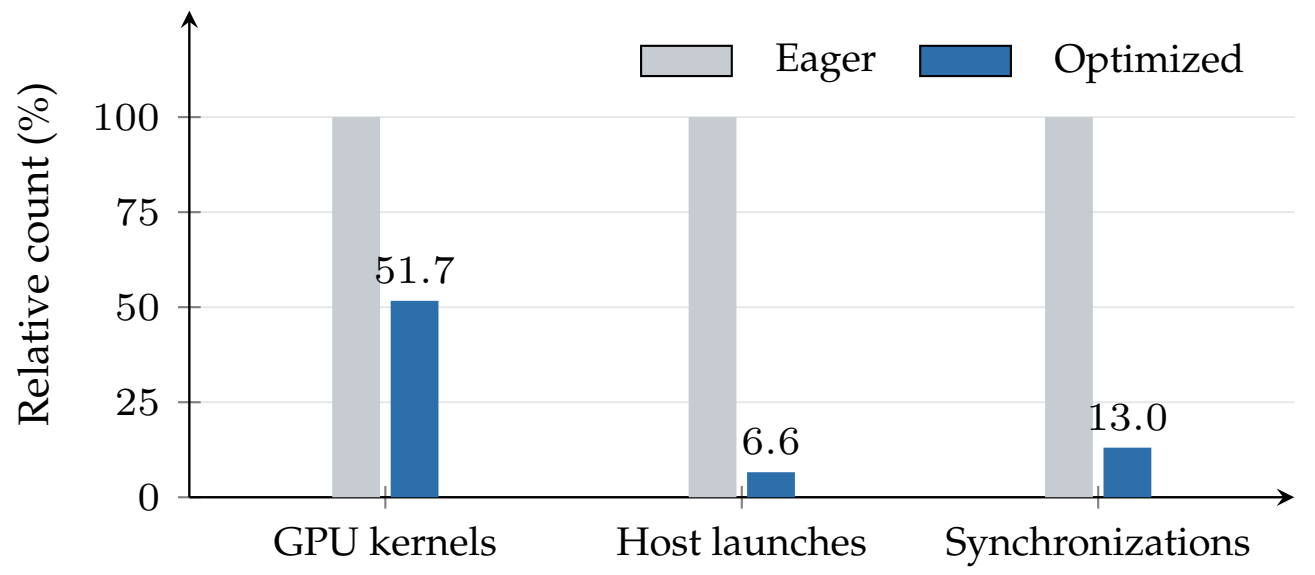

\resultheading{Execution profile and consistency}
Fig.~\ref{fig:pcm-profile} shows that ordinary host launches decrease from 83,777 to 5,519, a 93.4\% reduction, while GPU kernels decrease from 86,849 to 44,877, a 48.3\% reduction. Synchronization calls decrease from 10,194 to 1,330. The larger reduction in host submissions is consistent with graph replay reducing dispatch overhead for many small kernels. The optimized path also preserves bitwise-identical generated outputs and committed states across four 100-window rollouts with different seeds.

\begin{table}[!tb]
\centering
\caption{\textbf{Inference Optimization Component Ablation for PCM.} Flow, Fusion, and Graph denote execution-flow simplification, fused Mamba A/DT transition preprocessing, and CUDA Graph replay, respectively. Latencies are in ms ($\downarrow$); speedup is the median paired ratio ($\uparrow$).}
\label{tab:pcm-ablation}
\begingroup
\footnotesize
\setlength{\tabcolsep}{2.3pt}
\renewcommand{\arraystretch}{1.13}
\begin{tabular*}{\linewidth}{@{\extracolsep{\fill}}cccrrr@{}}
\toprule
Flow & Fusion & Graph & p50 & p95 & Speedup \\
\midrule
--- & --- & --- & 1685.35 & 2471.65 & $1.000\times$ \\
$\checkmark$ & --- & --- & 1102.82 & 1647.52 & $1.525\times$ \\
$\checkmark$ & $\checkmark$ & --- & 1069.52 & 1588.74 & $1.578\times$ \\
$\checkmark$ & --- & $\checkmark$ & 285.20 & 352.22 & $5.943\times$ \\
$\checkmark$ & $\checkmark$ & $\checkmark$ & \textbf{276.14} & \textbf{345.82} & $\mathbf{6.138\times}$ \\
\bottomrule
\end{tabular*}
\endgroup
\end{table}

\section{Real-World Evaluation}
\label{sec:exp5}

\begin{figure*}[tb]
\centering
\includegraphics[width=\linewidth,height=0.70\textheight,keepaspectratio]{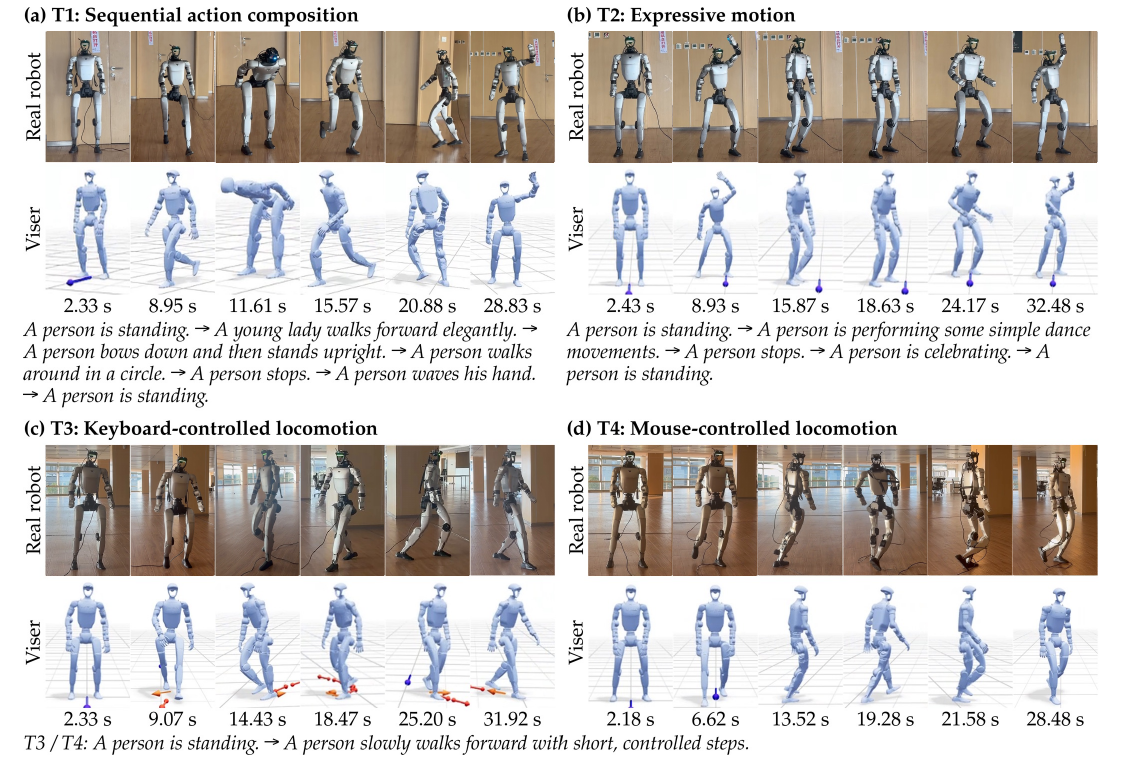}
\caption{\textbf{Real-world qualitative results on Unitree G1.} (a) T1, sequential action composition; (b) T2, expressive motion; (c) T3, keyboard-controlled locomotion; and (d) T4, mouse-controlled locomotion. Each panel shows real-robot frames, Viser references, timestamps, and the complete prompt sequence.}
\label{fig:exp5-g1}
\end{figure*}

While Sec.~\ref{sec:main-results} assesses long-horizon spatial constraints and command agreement in generated motion references, Sec.~\ref{sec:inference-efficiency} evaluates the cost of producing each motion window. We next examine whether these references support continuous and responsive execution on a physical Unitree G1. The experiments measure sequential text-guided action composition, online spatial control, and physical response in the integrated generation-and-tracking system. Fig.~\ref{fig:exp5-g1} presents representative robot executions alongside their Viser references.

\noindent\textbf{Evaluation setting.}
We conduct 20 trials each for sequential action composition (T1) and expressive motion (T2), and 30 trials each for keyboard-controlled (T3) and mouse-controlled (T4) locomotion. All conditions use the same motion generator, retargeting pipeline, and whole-body controller, with paired initial states, random seeds, and command sequences. Full-task success requires completion of all prescribed stages without a fall, protective stop, or human intervention. We additionally evaluate stage completion, command transitions, tracking accuracy, stance slip, and response latency. The model generates 52-frame windows at 25\,Hz with four windows of recent context. Detailed evaluation criteria are provided in App.~D.

\noindent\textbf{Deployment architecture.}
The deployment comprises a remote RTX 4090 server for text encoding and motion generation, an intermediate workstation for communication and interaction, and the onboard G1 controller for motion tracking. Generated motion references are transmitted through the workstation via ZeroMQ and SSH tunneling at a 50-Hz transport rate, distinct from the model's 25-Hz motion representation. The SONIC controller tracks the received references without executing the motion generator onboard.

\noindent\textbf{Interactive timing.} 
We distinguish handoff latency from physical-response latency. Handoff latency is measured from system receipt of a text or spatial command to completion of the
corresponding motion reference on the GPU server. It includes motion generation, decoding, accepted-state updates, and text encoding when a new text instruction requires it, while excluding network transmission and robot-side processing. Physical-response latency starts at the same command receipt event and ends at the onset of sustained motion satisfying the command tolerance on the G1, additionally accounting for communication, buffering, and physical execution delays. These measurements are distinct from the cached-text inference timings in Sec.~\ref{sec:inference-efficiency}.

\subsection{Motion Tracking with SONIC}
\label{sec:sonic-demo}
\begin{figure*}[tb]
\centering
\includegraphics[width=\linewidth,height=0.70\textheight,keepaspectratio]{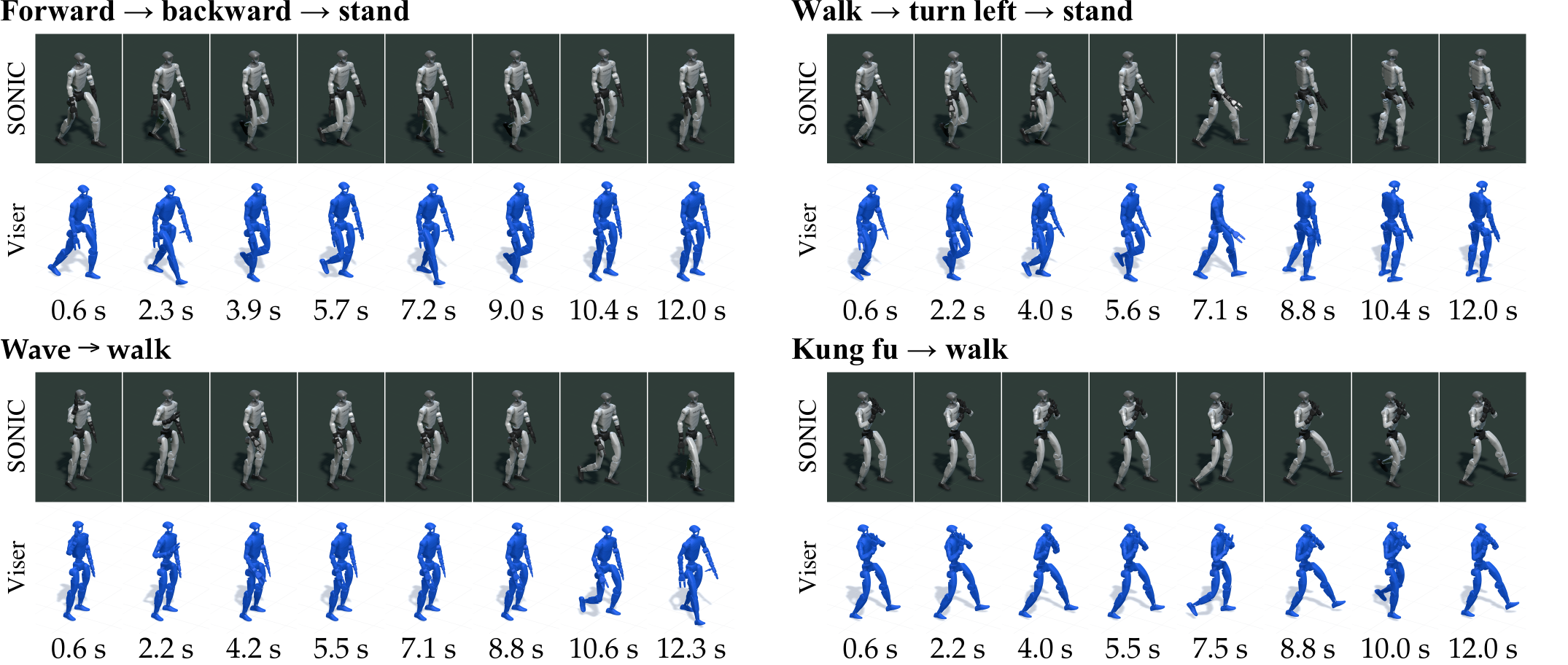}
\caption{\textbf{Simulated motion tracking with SONIC.} Top left: forward to backward walking; top right: walking with a left turn; bottom left: waving followed by walking; bottom right: martial-arts motion followed by walking. Each quadrant compares simulated execution (upper row) with the corresponding Viser reference (lower row) at matching timestamps.}
\label{fig:sonic-interaction}
\end{figure*}

We use SONIC~\cite{SONIC} to track the generated motion references in MuJoCo. Fig.~\ref{fig:sonic-interaction} shows four instruction sequences, including locomotion reversal, directional turning, and transitions between articulated actions and walking. For each sequence, the upper row shows physics-based execution in MuJoCo, where the SONIC controller tracks the generated motion reference and the resulting simulated motion is rendered. The lower row shows a kinematic visualization of the same reference in Viser: forward kinematics is used to update the robot mesh, which is then rendered without controller tracking or physics simulation. The two rows are shown at matching timestamps to compare the simulated execution with the reference motion. These examples qualitatively illustrate the transfer of generated motion sequences to a tracking controller under changing instructions, complementing the quantitative physical evaluation below.

\subsection{Sequential and Expressive Motion}
\label{sec:exp5-composition}

The T1 example contains changes in both global displacement and articulated pose. Forward progression is followed by a visible bow at 11.61\,s and a raised-arm gesture at 28.83\,s, with intermediate changes in body orientation (Fig.~\ref{fig:exp5-g1}, panel (a)). T2 exhibits a different combination of arm elevation, torso orientation, and lower-body configuration, extending to the 32.48\,s sample (Fig.~\ref{fig:exp5-g1}, panel (b)). These examples show that the physical repertoire extends beyond a repeated walking cycle. The robot and reference preserve recognizable action structure, while differences in trunk inclination and foot placement remain visible. 

Tab.~\ref{tab:exp5-main} summarizes sequential action execution on the physical G1. Full-task success reaches 18/20 (90.0\%) for T1 and 19/20 (95.0\%) for T2, with 136/140 and 99/100 completed stages, respectively. Of the scheduled text transitions, 117/120 in T1 and 78/80 in T2 succeed. These results complement the long-horizon command-matching evaluation in Sec.~\ref{sec:main-results} by showing that successive generated motion references can be tracked through multiple text-guided stages on the robot.

\subsection{Interactive Control and Physical Stability}
\label{sec:exp5-control}
\unskip
Fig.~\ref{fig:exp5-g1}(c) and (d) demonstrate interactive locomotion through keyboard and mouse inputs, respectively.
T3 and T4 test live spatial commands together with changes in text instructions. Tab.~\ref{tab:exp5-main} reports full-task success of 29/30 (96.7\%) for keyboard control and 28/30 (93.3\%) for mouse control, with 30/30 and 29/30 successful scheduled text transitions, respectively. These results extend the evaluation of conditional motion generation to interactive commands during physical execution on the G1.

Tab.~\ref{tab:exp5-control} reports tracking accuracy and responsiveness under interactive control. Keyboard and mouse interfaces yield speed RMSEs of 0.041/0.045\,m/s, heading MAEs of $4.6^\circ$/$5.1^\circ$, and median physical-response latencies of 351/360\,ms, respectively. Within the 1.5-s response deadline, 296/300 (98.7\%) keyboard events and 292/300 (97.3\%) mouse events are successfully executed. Stance-slip velocities remain low at 1.5/1.7\,cm/s, reflecting limited foot sliding during interactive locomotion.

\begin{table*}[!tb]
\centering
\caption{\textbf{Task Completion and Execution Stability.} Stable denotes execution without a fall, protective stop, or human intervention. Text switches measure successful transitions between scheduled text instructions across
all four tasks, while spatial-control events are evaluated separately.}
\label{tab:exp5-main}
\begingroup\fontsize{8.4}{10}\selectfont
\begin{tabular*}{\linewidth}{@{\extracolsep{\fill}}lrrrrrr@{}}
\toprule
Task & Full success \up & SR (\%) [95\% CI] \up & Stages \up & Text switches \up & Stable \up & Falls \dn \\
\midrule
T1: action composition & 18/20 & 90.0 [69.9, 97.2] & 136/140 & 117/120 & 19/20 & 0 \\
T2: expressive motion & 19/20 & 95.0 [76.4, 99.1] & 99/100 & 78/80 & 20/20 & 0 \\
T3: keyboard control & 29/30 & 96.7 [83.3, 99.4] & 59/60 & 30/30 & 29/30 & 0 \\
T4: mouse control & 28/30 & 93.3 [78.7, 98.2] & 58/60 & 29/30 & 29/30 & 0 \\
\bottomrule
\end{tabular*}\endgroup
\end{table*}

Across all four tasks, 94/100 trials achieve full-task
success, 97/100 meet the stability criterion, and no
falls are observed (Tab.~\ref{tab:exp5-main}).
Three trials remain stable without completing all
task requirements, while three fail the stability
criterion. Stance slip further characterizes contact
behavior. These results reflect the complete
generation-and-tracking system rather than
individual component failures.

\begin{table}[!tb]
\centering
\caption{\textbf{Interactive Control Accuracy and Response Latency.} Tracking and slip are episode mean $\pm$ SD; latency is p50/p95. Each interface has 30 episodes. The deadline counts include all 300 scheduled spatial events per interface.}
\label{tab:exp5-control}
\begingroup\fontsize{8.1}{9.7}\selectfont
\begin{tabular*}{\linewidth}{@{\extracolsep{\fill}}lrr@{}}
\toprule
Measure & Keyboard & Mouse \\
\midrule
Speed RMSE (m/s) \dn & $0.041\pm0.012$ & $0.045\pm0.014$ \\
Heading MAE (deg) \dn & $4.6\pm1.5$ & $5.1\pm1.7$ \\
Handoff p50/p95 (ms) \dn & 246 / 302 & 251 / 310 \\
Response p50/p95 (ms) \dn & 351 / 504 & 360 / 525 \\
Stance slip (cm/s) \dn & $1.5\pm0.5$ & $1.7\pm0.6$ \\
Events within 1.5\,s \up & 296/300 & 292/300 \\
\bottomrule
\end{tabular*}\endgroup
\end{table}

\section{Limitations and Future Work}
Our current framework focuses on controllable humanoid motion generation and does not yet address object manipulation. Future work will focus on whole-body loco-manipulation, coordinating locomotion with object manipulation.

\section{Conclusion}
\label{sec:conclusion}
We presented Motion Mamba 2.0, an autoregressive diffusion framework
that combines local attention, resumable Mamba states, and persistent
content memory. Evaluations across three G1 motion benchmarks demonstrate
improved terminal constraint accuracy, achieving Root CE of 7.55\,cm
and End CE of 13.64\,cm at 39.52\,s on retargeted HiPHI,
corresponding to the lowest Root CE and End CE among the evaluated methods.
Architectural ablations, including text-conditioning and state-difference variants, support the contributions of recurrent modeling and PCM to extended spatial control.
The proposed execution optimizations further reduce median complete-window
latency to 276.14\,ms on an RTX 4090, achieving a median paired speedup
of $6.138\times$ while preserving the tested outputs and committed states.
Physical G1 experiments achieve full-task completion in 94 of 100 trials
and demonstrate multi-stage text-guided execution and responsive
spatial control through the integrated generation-and-tracking system.
These results demonstrate the effectiveness of our framework in generating
long-horizon, real-time controllable humanoid motion with bounded historical context.

\section*{Acknowledgement}
This work was supported by the National Natural Science Foundation of China (NSFC) under Grant No. 62676007.

\bibliographystyle{IEEEtran}
\bibliography{reference}

\begin{IEEEbiography}[{\includegraphics[width=1in]{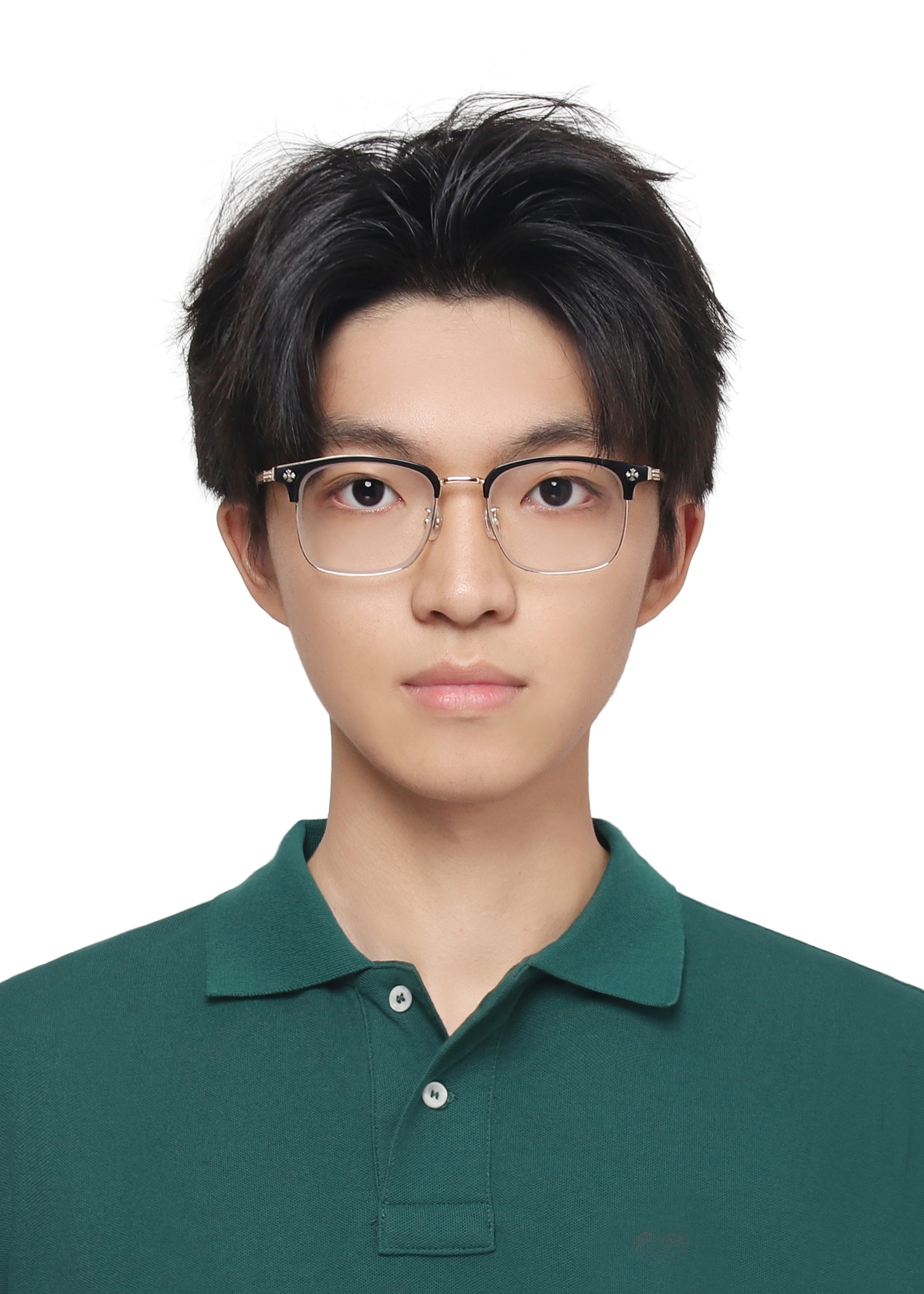}}]{Yue Huang}
is an undergraduate student at South China University of Technology. Currently, his main research interests lie in Humanoid locomotion, 3D perspective modeling, world models, and quantitative investment.
\end{IEEEbiography}

\begin{IEEEbiography}[{\includegraphics[width=1in]{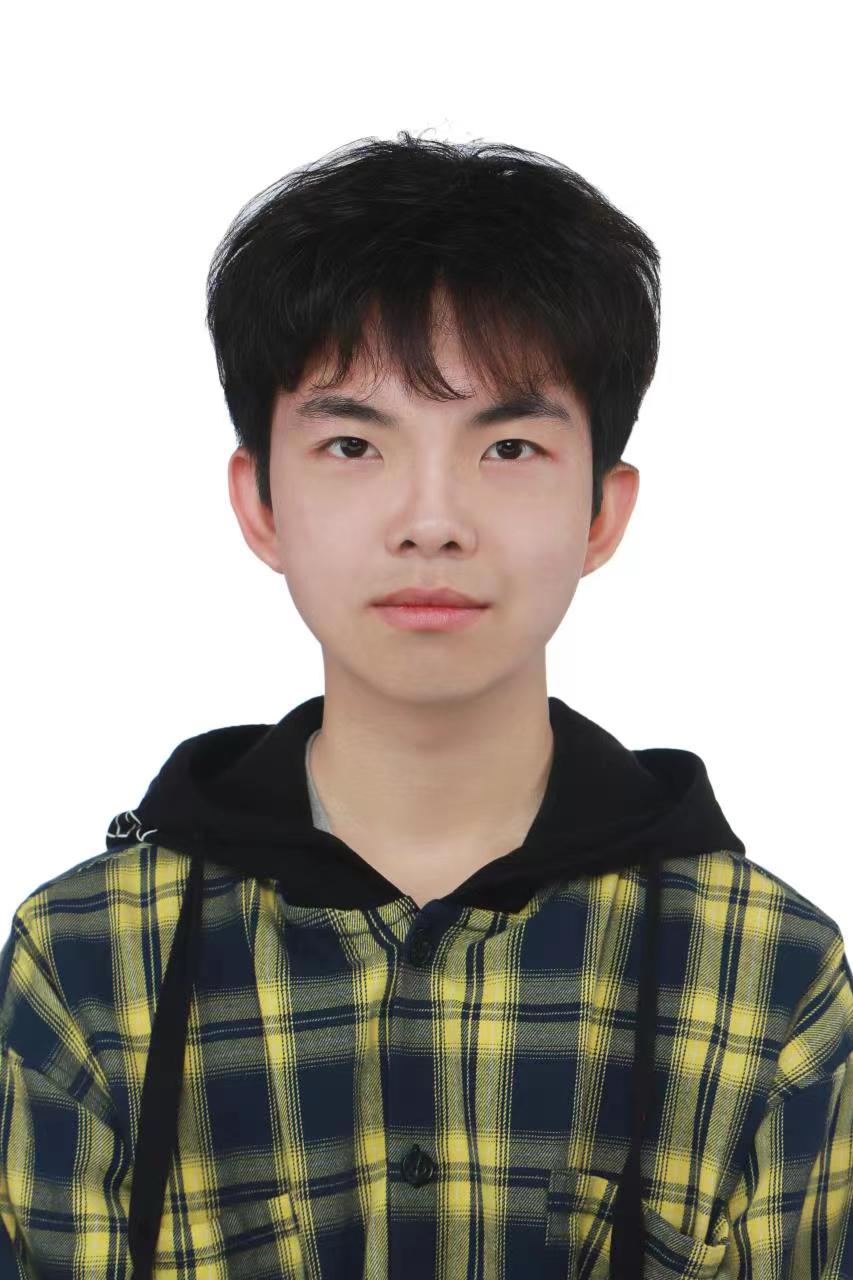}}]{Tianshan Zhang}
is currently a research assistant at Peking University. Previously, he was a research intern at the Institute of Automation, Chinese Academy of Sciences (CASIA). His research interests include embodied AI and robot learning, with a particular focus on generative models, vision-language-action models, and reinforcement learning.
\end{IEEEbiography}

\begin{IEEEbiography}[{\includegraphics[width=1in]{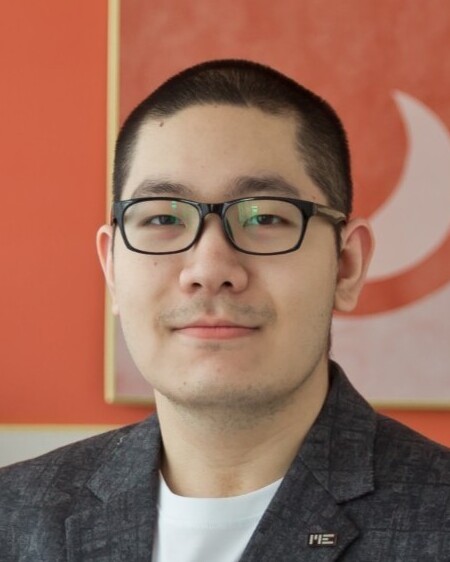}}]{Zeyu Zhang}
is a researcher working on generative AI, with a particular interest in building models that understand and interact with the physical world. He received his bachelor’s degree from the Australian National University, where he was advised by Prof. Richard Hartley and Prof. Ian Reid. His research explores generative modeling for learning physical dynamics from visual data. His work spans world models, multimodal foundation models, embodied AI, and AI for health. 
\end{IEEEbiography}

\begin{IEEEbiography}[{\includegraphics[width=1in]
{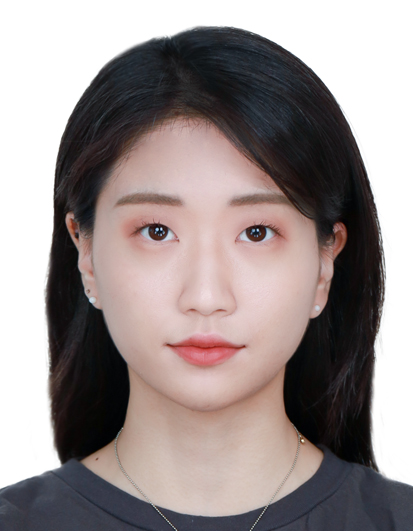}}]{Jayeon Kang} is currently pursuing an M.S. degree in artificial intelligence at Korea University, South Korea.  Her research interests span 3D computer vision, human motion generation and generative modeling. She is also interested in state-space models and their potential for embedded AI.
\end{IEEEbiography}

\begin{IEEEbiography}[{\includegraphics[width=1in,height=1.25in,clip,keepaspectratio]{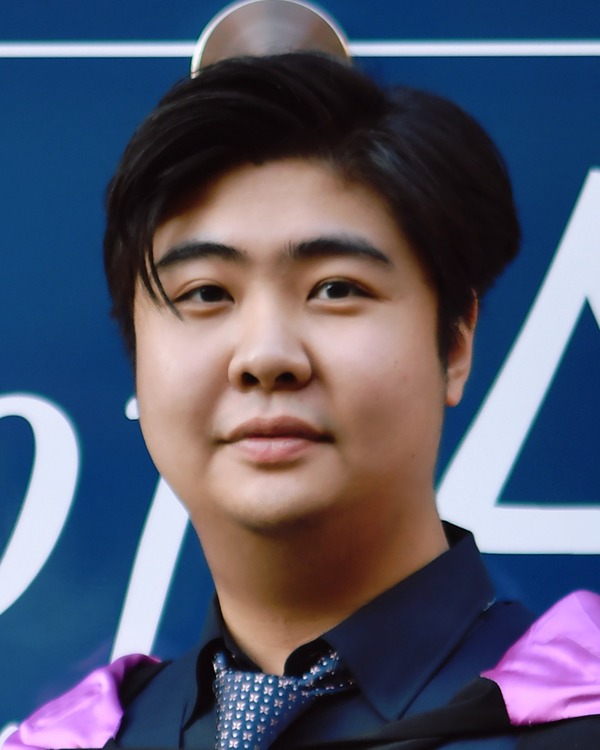}}]{Akide Liu}
is currently pursuing the Ph.D. degree in computer science at Monash University, Australia, under the supervision of Dr. Bohan Zhuang and Prof. Gholamreza Haffari. Akide received the B.S. (Honours) degree in computer science with First Class Honours from the University of Adelaide, Australia, in 2023, and has been a research intern at ByteDance and at DAMO Academy, Alibaba Group. Research interests include efficient AI, unified vision-language models, and generative models.
\end{IEEEbiography}

\begin{IEEEbiography}[{\includegraphics[width=1in,height=1.25in,clip,keepaspectratio]{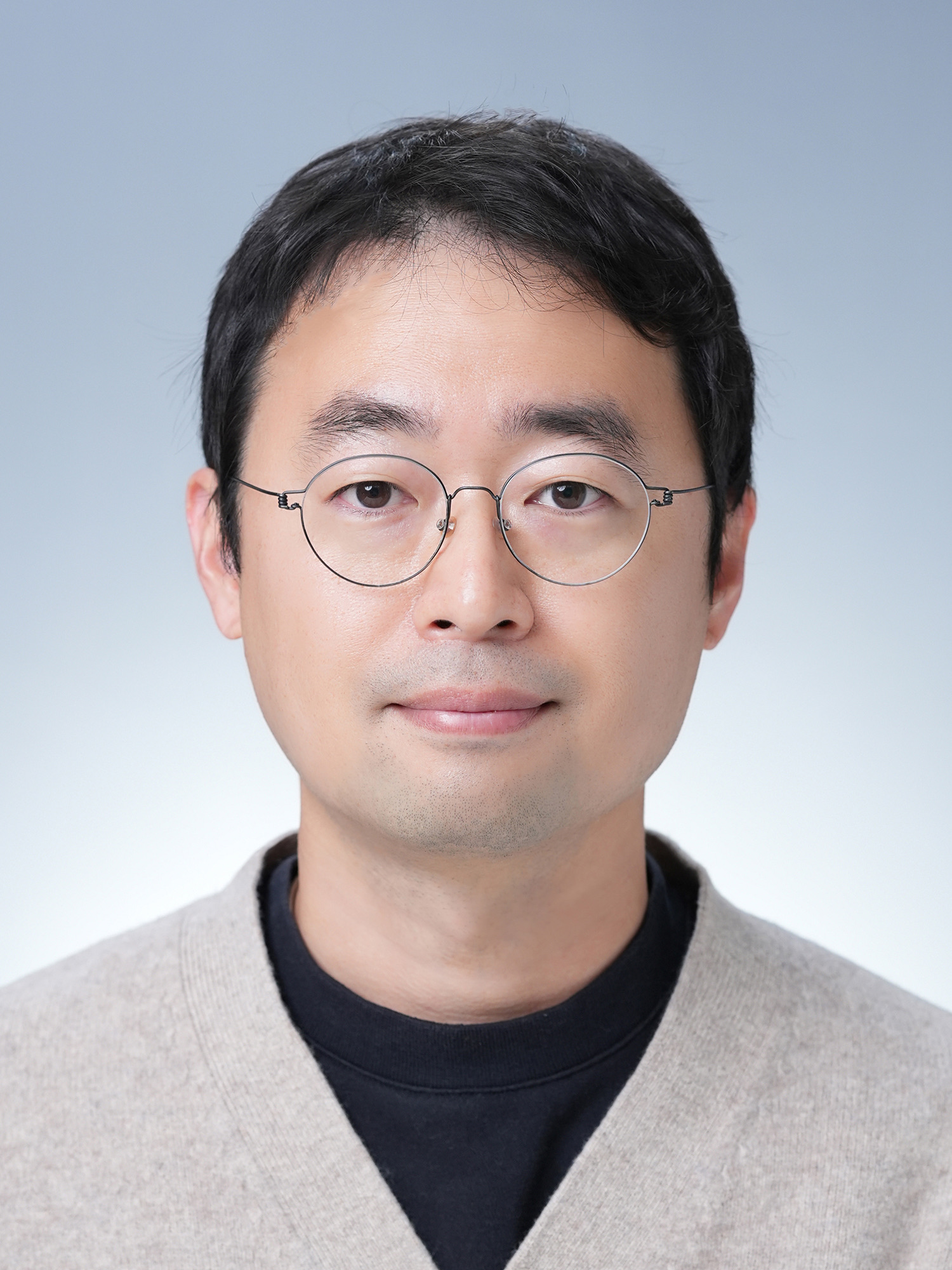}}]{Sungjoon Choi}
is an Associate Professor in the Department of Artificial Intelligence at Korea University, South Korea. Previously, he was a Staff Software Engineer at Google Research in Mountain View, CA, USA. He received the Ph.D. degree in electrical engineering from Stanford University under the supervision of Vladlen Koltun and the B.S. degree in computer science and engineering from Seoul National University. His research interests include generative models, computational imaging, efficient visual computing, and 3D vision. He was a recipient of the Samsung Scholarship and a two-time finalist in the ACM-ICPC World Finals.
\end{IEEEbiography}

\begin{IEEEbiography}[{\includegraphics[width=1in,height=1.25in,clip,keepaspectratio]{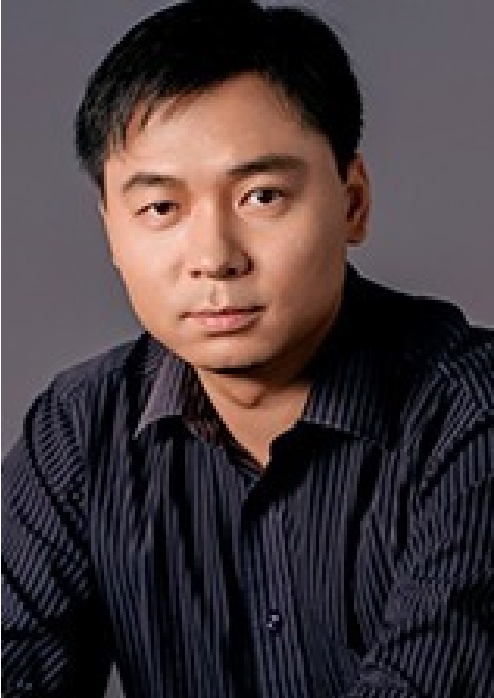}}]{Dacheng Tao} (Fellow, IEEE)
is currently a Distinguished University
Professor in the College of Computing \& Data
Science at Nanyang Technological University. He
mainly applies statistics and mathematics to artificial
intelligence and data science, and his research is
detailed in one monograph and over 200 publications
in prestigious journals and proceedings at leading
conferences, with best paper awards, best student
paper awards, and test-of-time awards.  He received the
2015 and 2020 Australian Eureka Prize, the 2018 IEEE ICDM Research
Contributions Award, and the 2021 IEEE Computer Society McCluskey
Technical Achievement Award. He is a Fellow of the Australian Academy of Science, AAAS, ACM and IEEE.
\end{IEEEbiography}

\begin{IEEEbiography}[{\includegraphics[width=1in,height=1.25in,clip,keepaspectratio]{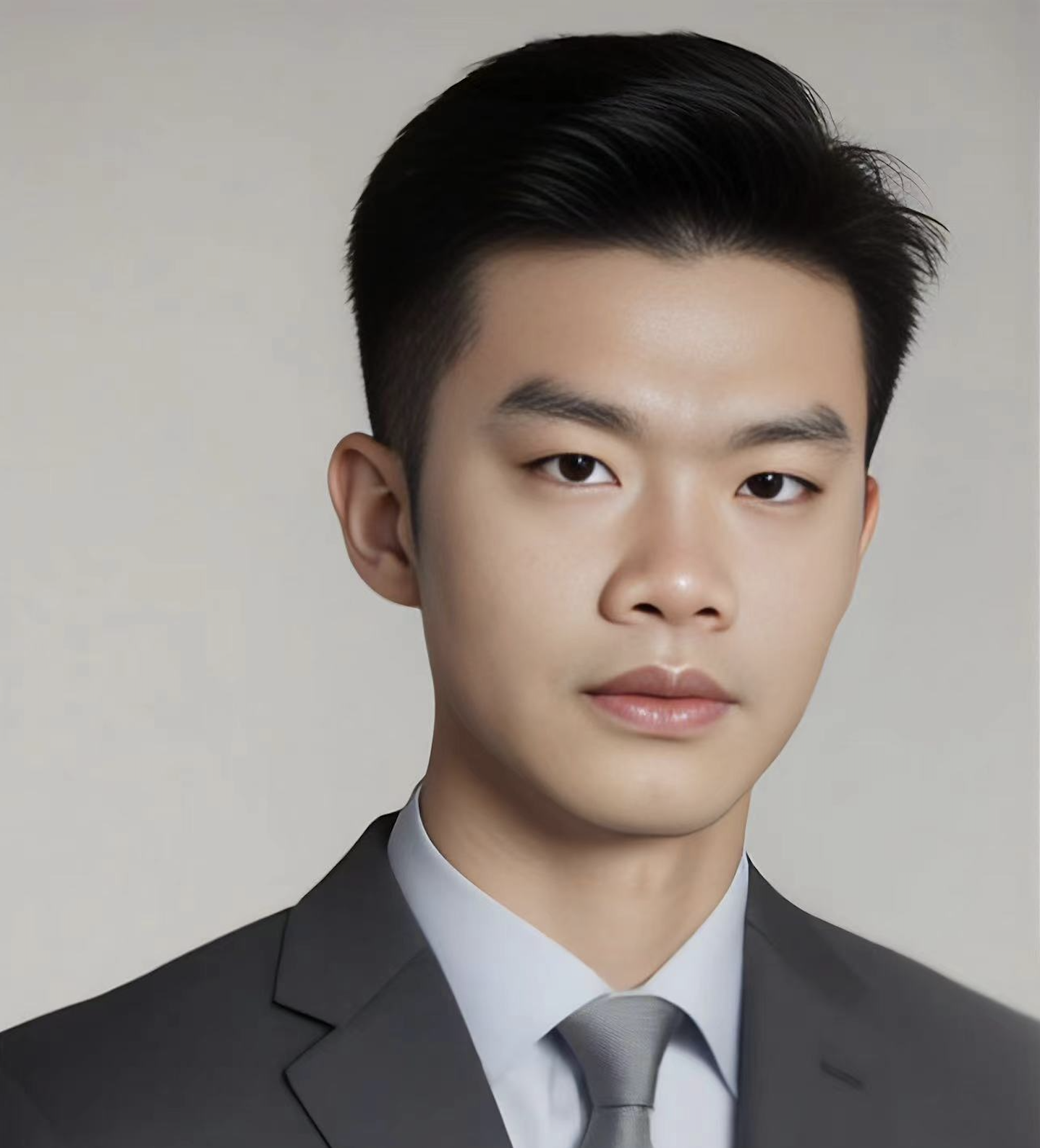}}]{Hao Tang}
is an Assistant Professor at Peking University, China. Previously, he held postdoctoral positions at CMU, USA, and ETH Z\"{u}rich, Switzerland. He earned his master's degree from Peking University, and his Ph.D. from the University of Trento, Italy.
He has had the opportunity to visit the University of Oxford, NEU, NUS, and IIAI, among other institutions.
His research focuses on generative AI, world models, spatial intelligence, embodied AI, LLMs, MLLMs, and computer vision.
\end{IEEEbiography}

\flushcolsend
\clearpage
\appendices
\section{Dataset Construction and Retargeting}
\label{app:implementation}

\subsection{Corpus and Conversion}

Tab.~\ref{tab:appendix_splits} reports the fixed splits. Source families and admitted mirrors follow the corresponding split manifests; the reported durations count stored frames at 25\,Hz. HumanML3D motions are reconstructed from AMASS/SMPL-H parameters, and the selected high-precision human-motion subset of HiPHI uses BVH forward kinematics. The HiPHI object-interaction subset is excluded. BONES Seed retains its official G1 trajectories.

We use GMR with the same 29-DoF G1 kinematic model for retargeting and geometric evaluation. Source-specific rest-pose transforms and pelvis-height scaling align the human skeletons with G1. Root translations and local rotations are resampled to 25\,Hz using linear interpolation and SLERP. The solver combines position and orientation objectives under joint limits, followed by temporal arm refinement, collision refinement, and bounded ground and rate corrections. Forward-kinematic consistency of the robot assets is verified on 123 configurations.

\begin{table*}[!b]
\centering
\caption{\textbf{Frozen corpus composition.} Training, validation, and test columns count stored records. Durations include admitted mirrors and are computed at 25\,Hz.}
\label{tab:appendix_splits}
\footnotesize
\begin{tabular*}{\linewidth}{@{\extracolsep{\fill}}lrrrrr@{}}
\toprule
Source & Training & Validation & Test & Training hours & All-split hours \\
\midrule
BONES Seed & 124,658 & 7,578 & 9,384 & 256.24 & 286.95 \\
HumanML3D & 14,612 & 948 & 2,965 & 28.50 & 36.37 \\
HiPHI & 17,549 & 1,472 & 482 & 304.18 & 337.67 \\
\midrule
Total & 156,819 & 9,998 & 12,831 & 588.92 & 660.98 \\
\bottomrule
\end{tabular*}
\end{table*}

\subsection{Mapped-Target Residuals and Representation Checks}

The assessment covers all 18,525 HumanML3D and 19,503 HiPHI records across the three splits. Let $\mathbf p^{\mathrm{G1}}_{itj}$ be the robot body-origin position and $\widetilde{\mathbf p}^{\mathrm H}_{itj}$ its scaled, offset human target, in meters. For clip $i$ and body group $b$, the per-clip residual in centimeters is
\begin{equation}
e_i^{(b)}=\max_{j\in\mathcal J_b}Q_{0.95,t}
\left(100\left\|\mathbf p^{\mathrm{G1}}_{itj}-\widetilde{\mathbf p}^{\mathrm H}_{itj}\right\|_2\right),
\label{eq:retarget_residual}
\end{equation}
where $Q_{0.95,t}$ is the 95th percentile over valid frames, and $\mathcal J_b$ contains the pelvis or both ankles. Tab.~\ref{tab:appendix_residuals} summarizes these values with equal clip weights.

\begin{table}[!t]
\centering
\caption{\textbf{Mapped-target residuals in centimeters.} Median, p95, and maximum summarize the per-clip residual defined above.}
\label{tab:appendix_residuals}
\footnotesize
\begin{tabular*}{\linewidth}{@{\extracolsep{\fill}}llrrr@{}}
\toprule
Source & Body & Median & p95 & Max. \\
\midrule
HumanML3D & Pelvis & 0.99 & 3.07 & 9.65 \\
 & Ankle & 1.49 & 3.37 & 10.53 \\
\addlinespace
HiPHI & Pelvis & 0.56 & 1.04 & 11.25 \\
 & Ankle & 0.93 & 2.39 & 29.06 \\
\bottomrule
\end{tabular*}
\end{table}

Final rate projection leaves joint angles unchanged in 37,881/38,028 records (99.61\%). The largest final angle corrections are 0.254 and 0.229\,rad for HumanML3D and HiPHI; the largest root-height lifts are 10.85 and 10.55\,cm. All admitted human-source records pass the stored-format and 414-D round-trip checks, with maximum position discrepancy $1.31\times10^{-6}$\,m. These checks establish numerical and representation consistency; the mapped-target statistics quantify geometric alignment to the specified pelvis and ankle landmarks.

\section{Model and Inference Implementation}
\label{app:method}

\subsection{Layer State and PCM Updates}

Root and body layers use condition-modulated LayerNorm before the sequence mixer and feed-forward sublayer. Attention and Mamba receive the same mixer features. The Mamba state carries the accumulated phase, recurrent matrix, preceding rotated state-input vector, and preceding value vector. Phase and matrix accumulation use FP32; the latter vectors follow the BF16 scan boundary. A session reset clears these tensors, the K/V cache, PCM slots, and preceding summaries together.

The PCM projector averages the recurrent matrix over the state dimension, flattens head and value axes, and applies a linear map and LayerNorm. Successive accepted summaries define the state difference. The masked mean of valid text embeddings supplies the write condition, independently of diffusion time and heading. A linear projection and LayerNorm form the text features; a linear conditioner produces scale and shift parameters for the normalized state difference. The summary, modulated difference, and text features construct the candidate content in Eq.~(11). A linear decoder and featurewise LayerNorm produce slot proposals, while a linear sigmoid writer also reads the previous slots to predict the write gate in Eq.~(12). The write-rate cap is $\eta=0.25$. Gates are broadcast over features, and slots are updated once per accepted window. During denoising, all guidance branches read the same committed cache, Mamba state, slots, and summary.

\noindent\textbf{Training-time context.}
The complete denoiser, including PCM, is present from the beginning of training and is optimized under the clean-window generation objective; no separate PCM initialization or fine-tuning stage is used. Each training example predicts a single 52-frame generation window conditioned on sampled clean ground-truth historical motion, rather than on an autoregressively generated prefix. Training does not unroll multiple generated windows or perform cross-window truncated backpropagation through time. During inference, by contrast, persistent states are updated from accepted generated windows. The zero-initialized PCM readout scale in Eq.~(13) suppresses the additional attention contribution at initialization; it does not imply that PCM is inserted after a separately trained Native stage.
During joint training, the root-to-local conditioning path is detached before body prediction. Body losses therefore do not
backpropagate through this conversion into the root denoiser, which remains supervised by its own prediction losses.

\subsection{Accepted-History Streaming Inference}

The generation algorithm below makes root--body prediction, guidance, and persistent updates explicit. The clean pass uses the same root-to-local transformation and positional indexing as denoising. Future control tokens do not enter persistent history. Cache projections are refreshed whenever their model parameters change. In the algorithm, $\kappa$ indexes the three guidance branches, $\boldsymbol\nu$ denotes accepted-token validity, and $\mathcal S$ contains the complete resumable Mamba state; $\mathbf H$ is its recurrent matrix. The branch loop specifies logical dependencies; the optimized implementation captures the three guidance branches together within each of the separate root and body DiT-stack graphs. Layer indices are omitted inside the accepted-state update loop for compactness; both denoisers retain separate layerwise states. Temporary denoising states are discarded. The subscript $\kappa$ on input construction applies the corresponding text and spatial condition masks. $\operatorname{TextContext}^{q,\ell}$ denotes the text projection, difference modulation, and concatenation in Eqs.~(9)--(11).

\begin{algorithm}[!tb]
\caption{Diffusion-Safe Streaming Generation}
\label{alg:generation}
\footnotesize
\begin{algorithmic}[1]
\Require Conditions $\{s_t,\mathbf g_t,\mathbf v_t\}$ for $N$ windows, initial history $\mathbf m_{\rm obs}$,
\Statex \hspace{\algorithmicindent} diffusion steps $K$, cache width $W$, scales $(w_s,w_g)$
\State $\memory_{-1}\gets\operatorname{Init}(\mathbf m_{\rm obs})$; $\mathbf T\gets\mathbf0$
\For{$t=0,\ldots,N-1$}
  \State $\memory^{\rm read}\gets\memory_{t-1}$; $\mathbf e_t\gets\operatorname{PoolText}(s_t)$
  \State $\mathbf z_{t,K}\sim\mathcal N(\mathbf0,\mathbf I)$
  \For{$\tau=K,\ldots,1$}
    \For{$\kappa\in\{s,g,\varnothing\}$}
      \State $\mathbf c_\kappa\gets\operatorname{Cond}_\kappa(s_t,\mathbf g_t,\mathbf v_t,\tau)$
      \State $\mathbf u^r_\kappa\gets\operatorname{RootInput}_\kappa(\mathbf z_{t,\tau},\mathbf g_t,\mathbf v_t)$
      \State $\widehat{\mathbf r}_\kappa\gets\mathcal F^r_\theta(\mathbf u^r_\kappa,\mathbf c_\kappa;\memory^{r,\rm read})$
      \State $\mathbf u^b_\kappa\gets\operatorname{BodyInput}_\kappa(\operatorname{Local}(\widehat{\mathbf r}_\kappa),\mathbf z_{t,\tau},\mathbf g_t,\mathbf v_t)$
      \State $\widehat{\mathbf b}_\kappa\gets\mathcal F^b_\theta(\mathbf u^b_\kappa,\mathbf c_\kappa;\memory^{b,\rm read})$
      \State $\widehat{\mathbf x}_\kappa\gets[\widehat{\mathbf r}_\kappa;\widehat{\mathbf b}_\kappa]$
    \EndFor
    \State $\widehat{\mathbf x}_0\gets\widehat{\mathbf x}_\varnothing+w_s(\widehat{\mathbf x}_s-\widehat{\mathbf x}_\varnothing)$
    \Statex \hspace{3em} $+\,w_g(\widehat{\mathbf x}_g-\widehat{\mathbf x}_\varnothing)$
    \State $\mathbf z_{t,\tau-1}\gets\mathcal S_\tau(\mathbf z_{t,\tau},\widehat{\mathbf x}_0)$
  \EndFor
  \State $(\mathbf x'_t,\mathbf o_t)\gets\operatorname{Recenter}(\mathbf z_{t,0})$
  \State $\{\mathbf K,\mathbf V,\boldsymbol\nu,\widetilde{\mathcal S}\}_{q,\ell}\gets\operatorname{Canonical}_{r\to b}(\mathbf x'_t;\memory^{\rm read})$
  \ForAll{root/body layers $(q,\ell)$}
    \State $\widetilde{\mathcal C}\gets\operatorname{FIFO}_W(\mathcal C_{t-1}\cup\{\mathbf K,\mathbf V,\boldsymbol\nu\})$
    \State $\overline{\mathbf s}\gets\Pi(\widetilde{\mathbf H})$; $\mathbf d\gets\overline{\mathbf s}-\overline{\mathbf s}_{t-1}$
    \State $\mathbf j\gets\operatorname{TextContext}(\overline{\mathbf s},\mathbf d,\mathbf e_t)$
    \State $\mathbf Z\gets\operatorname{LN}(f_p(\mathbf j))$
    \State $\mathbf w\gets\eta\sigma(f_w([\mathbf j;\operatorname{vec}(\mathbf P_{t-1})]))$
    \State $\widetilde{\mathbf P}\gets(1-\mathbf w)\odot\mathbf P_{t-1}+\mathbf w\odot\mathbf Z$
    \State $\pending^{q,\ell}\gets\{\widetilde{\mathcal C},\widetilde{\mathcal S},\widetilde{\mathbf P},\overline{\mathbf s}\}$
  \EndFor
  \If{$\operatorname{Valid}(\pending)$}
    \State $\memory_t\gets\operatorname{AtomicCommit}(\pending)$; $\mathbf T\gets\mathbf T+\mathbf o_t$
    \State $\operatorname{Emit}(\operatorname{Global}(\operatorname{Decode}(\mathbf x'_t),\mathbf T))$
  \Else
    \State $\memory_t\gets\memory_{t-1}$
  \EndIf
\EndFor
\end{algorithmic}
\end{algorithm}

\subsection{Fusion, Guidance, and Precision}

For transition inputs $a_i,d_i$, decay floor $a_{\min}>0$, and step-size bias $b_\Delta$, preprocessing is
\begin{equation}
\begin{aligned}
\chi(a)&=\max(a,0)+(1-\min(a,0))^{-1},\\
A_i&=\min(-\chi(a_i),-a_{\min}),\\
\Delta_i&=\operatorname{softplus}(d_i+b_\Delta),
\qquad \Xi_i=\Delta_iA_i.
\end{aligned}
\label{eq:execution_preprocess}
\end{equation}
The fused A/DT kernel emits $\Delta_i$ and $\Xi_i$ in the original scan layout, keeping intermediate preprocessing in FP32 registers. The scan and attention operators are retained. Fused scale--shift and gated-residual operations preserve the reference casts and arithmetic ordering; fused multiply--add contraction is disabled where required for bitwise equivalence. Reported equivalence runs use FP32 attention, BF16 Mamba compute boundaries, FP32 state accumulation, and disabled TF32.

Three-branch classifier-free guidance combines text-only, spatial-only, and unconditional predictions:
\begin{equation}
\begin{aligned}
\widehat{\mathbf x}_0
&=\widehat{\mathbf x}_{0,\varnothing}
+w_s(\widehat{\mathbf x}_{0,s}-\widehat{\mathbf x}_{0,\varnothing})+w_g(\widehat{\mathbf x}_{0,g}-\widehat{\mathbf x}_{0,\varnothing}).
\end{aligned}
\label{eq:execution_cfg}
\end{equation}
The root and body DiT sequence stacks use separate CUDA Graphs. Within each stack, the text-only, spatial-only, and unconditional guidance branches execute serially and are captured together. Input projections, conditioning preparation, root-output processing, root-to-local conversion, and body-input construction remain outside the stack-level graphs. Branch predictions are combined according to the guidance equation above. Capture reduces host dispatch without changing guidance weights, model parameters, or the root-to-body dependency. Sampling, random-number handling, recentering, and accepted-state commitment also remain outside capture. Graphs are rebuilt when tensor layouts or parameter versions become incompatible; unsupported calls use the reference path.

\subsection{Complete-Window Timing and Equivalence}
\label{app:efficiency-protocol}

The fixed-checkpoint study follows the hardware, precision, sampling, and five-process paired protocol in Sec.~5.4. Complete-window timing includes state restoration, sampling, accepted-state commitment, decoding, and CPU snapshot/export; text encoding, loading, warmup, and capture are excluded. Speedup is the median paired ratio, with a 95\% interval from 10,000 process-block bootstrap resamples. Four seeded 100-window rollouts verify bitwise-identical outputs and committed states. This cohort is independent of the model-comparison timings.

\section{Benchmark Evaluation Protocols}
\label{app:evaluation-protocol}

\subsection{Common Evaluation Conditions}

All methods follow the paired protocol in
Sec.~5.1: 128 held-out initialization source families and three noise seeds per family yield 384 rollouts per duration. Initializations and root and end-effector goals are paired across methods. No future ground-truth motion is provided.

For HiPHI, the same 128 initialization families, initial motion histories, physical targets, and noise seeds are shared across horizons. These initialization families are distinct from the phase-specific motion-reference families and distinct-caption retrieval-query sets described below. Root CE and
End CE are averaged over all 384 terminal rollout measurements. Cmd.\ is computed as the mean frame-level coarse root-state agreement within each 52-frame window, averaged over windows within each rollout and then over all 384 rollouts. R@3 and
L2-FID follow the window-level aggregation described below.

Although initialization cases are paired across horizons, the instruction-phase composition and window weights differ. Cross-horizon differences therefore describe performance across the evaluated
horizons and do not isolate the effect of duration or temporal error accumulation.

Model-comparison latency uses two warmups and 32 paired batch-one rollouts with cached text, including generation, decoding, and accepted-state
updates. These timing samples are separate from the 384 quality-evaluation rollouts per duration.

\subsection{Training and Baseline Adaptations}
\label{app:training-configuration}

\noindent\textbf{Training configuration.}
Motion Mamba 2.0 is trained jointly with Mamba-3 and PCM enabled from initialization. The generator starts from randomly initialized parameters (initialization seed 123), rather than an official pretrained generator checkpoint; the shared G1 motion codec is frozen and text features are cached. The model is trained for 100,000 optimizer updates using AdamW with $(\beta_1,\beta_2)=(0.9,0.999)$, $\epsilon=10^{-8}$, weight decay $0.01$, and global gradient-norm clipping at $1.0$. The learning-rate schedule uses 500 warmup updates from $10^{-6}$ to $5\times10^{-5}$, followed by cosine decay to $10^{-5}$ over the 100,000-update schedule. The released training recipe uses global batch size 512 throughout all 100,000 optimizer updates, totaling 51.2 million sample exposures. There is no separate PCM pretraining or late-stage PCM fine-tuning.

Training samples draw from BONES Seed, retargeted HumanML3D, and retargeted HiPHI with probabilities $0.70$, $0.15$, and $0.15$. The generation window contains 52 frames. History/future-conditioning lengths $(H,F)$ take the values $(0,0)$, $(52,0)$, $(196,0)$, $(0,196)$, $(52,144)$, and $(144,52)$ with respective probabilities $0.10$, $0.25$, $0.20$, $0.10$, $0.20$, and $0.15$. Text dropout and spatial-constraint dropout are each $0.1$, and the probability of empty spatial conditioning is $0.2$. The supported spatial conditions are root trajectories, root waypoints, full keyframes, joint positions, and joint rotations. The largest sampled historical prefix is 196 frames (7.84\,s), shorter than the four-window 208-frame local cache. Thus, this training protocol does not directly supervise retrieval from model-generated histories beyond the local attention horizon; longer-horizon behavior is evaluated through autoregressive inference.

Diffusion training uses clean-window ($x_0$) prediction with a uniformly sampled base diffusion timestep and Gaussian forward noise. The base noise schedule is cosine, with offset $s=0.008$ and maximum $\beta=0.999$; the number of base diffusion steps is read from the loaded G1 configuration, not specified by the ten-step DDIM inference schedule. The four terms in Eq.~(15) have unit coefficients. The hybrid reconstruction term uses Smooth L1 with $\beta_{\mathrm{SmoothL1}}=1$ over the 148-dimensional hybrid representation; decoded-body and observed-goal terms use normalized L1, while forward-kinematics consistency measures joint-position disagreement in meters. The hybrid root/body subcomponents are diagnostic reductions, not separately weighted losses. These settings describe the common G1 diffusion/$x_0$ objective and should not be read as the objectives for every model family.

\noindent\textbf{Controlled baseline adaptations.}
All compared methods and architectural variants are independently trained on the same G1 corpus under 100,000 optimizer updates and the matched global batch-size schedule, using newly trained 100k checkpoints rather than their original released generator weights. Their objectives remain architecture-specific: the shared G1 diffusion adaptations use the clean-motion objective above, autoregressive action-code adaptations use token cross-entropy, and MoLingo retains its masked-flow velocity objective. These distinctions preserve the training formulation of each model family instead of imposing a diffusion objective on non-diffusion methods. The G1 adaptation and evaluation interface applies paired physical constraints and matched data splits to all compared methods.

Minimal uses FP32 PyTorch SDPA with CUTLASS memory-efficient attention and no fused-kernel or CUDA Graph acceleration. Native configurations use their admitted accelerated execution paths. Consequently, architectural tables primarily compare quality and control; their latency columns describe the complete evaluated implementations. The fixed-weight component study in Sec.~5.4 separately measures execution acceleration.

\subsection{BONES Seed and HumanML3D Retrieval and Reference Distribution}

BONES Seed and retargeted HumanML3D use separately trained,
frozen evaluators. For each query, R@3 ranks one text-matched
generated motion against 31 fixed real-motion distractors.
Within each benchmark, all methods share the same evaluator,
candidate construction rules, and sampling protocol.
BONES Seed retains its official G1 motion references.
The HumanML3D evaluator uses three fixed candidate pools
and two frozen retrieval heads. Its source pool contains
accepted motions shorter than 10\,s, limiting evaluation
to 2.08 and 8.32\,s. Under this protocol, R@3 measures
reference-anchored retrieval rather than retrieval among
32 generated motions.

\subsection{HiPHI Retrieval and Reference Distribution}

The HiPHI evaluator is independently trained and frozen for all compared generators. Initializations span the same 128 source families from three actors at every evaluated horizon. For instruction phases 1--4, respectively, the numbers of distinct-caption retrieval queries are 61/110/61/42, and the numbers of distinct reference source families are 93/128/93/63. These counts describe instruction phases, \emph{not} the four evaluation durations. Within each phase, all 128 initialization cases receive an assigned real-motion reference entry; repeated entries can originate from the same reference source family without excluding any generated rollouts. The detailed-instruction reference bank is separated by source family from generator training and evaluator training/development.

The numbers of windows allocated to phases 1--4 are $(0,0,0,1)$, $(1,1,1,1)$, $(3,1,2,3)$, and $(6,2,5,6)$ at 2.08, 8.32, 18.72, and 39.52\,s, respectively. In particular, the single-window evaluation uses phase 4. Initial motion histories, spatial goals, and noise seeds are paired across horizons, but phase composition, the number of windows per phase, and terminal deadlines differ. Cross-horizon metric changes therefore do not isolate the effect of duration alone.

For generated models, R@3 ranks 32 generated motion candidates from the same model, window, and noise seed. The phase-specific distinct-caption queries are evaluated with three fixed candidate pools, two frozen retrieval heads, and three generator noise seeds. R@3 is averaged within each window and then arithmetically averaged over windows at the corresponding horizon. Consequently, phase contributions are weighted by their window counts, not equally across the four phases; the resulting retrieval score is not based on 384 independent text queries.

All three benchmarks compute L2-FID from individually L2-normalized 256-dimensional motion features of phase-paired 52-frame windows. For HiPHI, each window compares 384 generated features (128 initializations by three noise seeds) with 128 assigned real-reference entries, one per initialization case. Multiple reference entries may share a source family, and their case-assignment multiplicities are retained. HiPHI L2-FID is the arithmetic mean of its window-level values, rather than FID over embeddings of complete 39.52-s trajectories. A real-reference distribution has zero L2-FID under self-comparison; the R@3 reference-calibration values are reported separately in Tab.~\ref{tab:reference-calibration}. Since evaluators, retrieval candidate constructions, and reference distributions differ, absolute R@3 and L2-FID scores are interpreted within each benchmark. Command matching measures coarse root-motion agreement rather than complete natural-language instruction fulfillment.

\subsection{Real-Motion Reference Calibration}

Tab.~\ref{tab:reference-calibration} reports real-motion
reference calibration scores under the three frozen
benchmark-specific evaluators. These scores characterize
reference-motion retrieval under the corresponding query
sets and candidate-construction protocols. They are not
generated-model results, teacher-model outputs, or
theoretical upper bounds.

For HiPHI, reference calibration uses phase-paired real motions and the corresponding frozen phase-specific queries. The four calibration values in Tab.~\ref{tab:reference-calibration} are arranged by evaluated horizon; their component query sets follow the phase-to-window allocation stated above. Generated models are evaluated separately using 32 same-window candidates produced by the same method and noise seed. For BONES Seed and HumanML3D, calibration uses paired real-motion candidates, whereas generated models use one matched generated motion and 31 fixed real-motion distractors.

Because the candidate constructions differ, the
reference scores are reported as evaluator diagnostics
rather than direct model-performance comparisons.

\subsection{Qualitative Long-Horizon Goal-Conditioning Comparison}
\label{app:goal-comparison}
\begin{figure*}[!t]
  \centering
  \includegraphics[width=\textwidth]{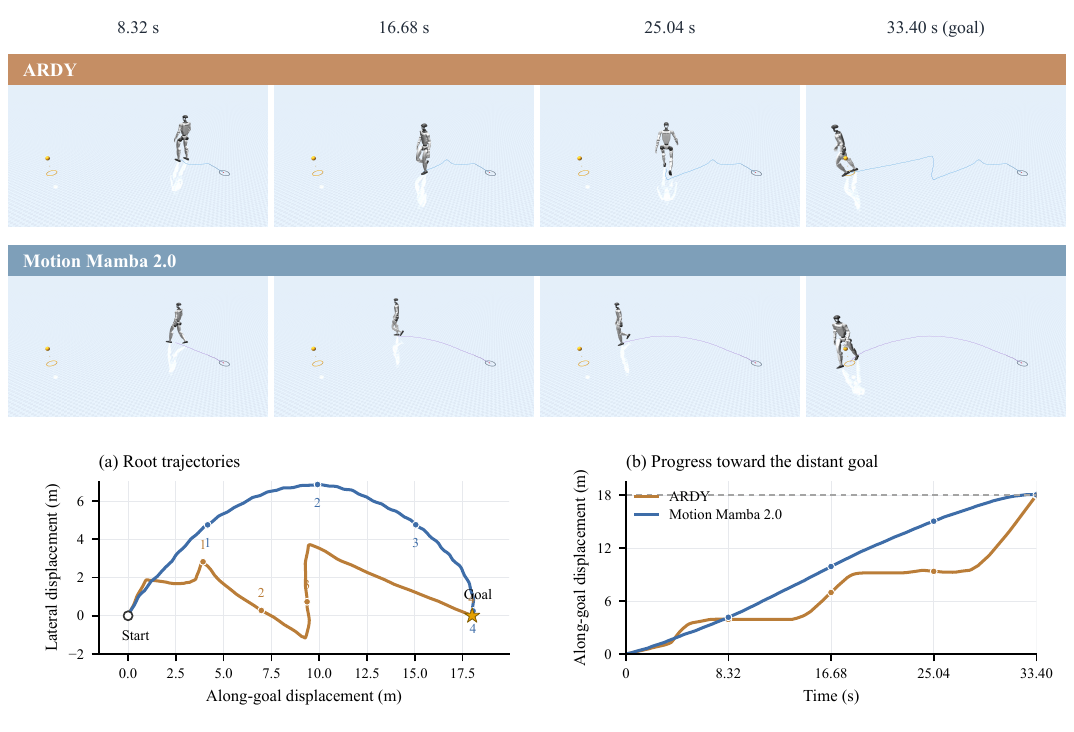}
  \caption{\textbf{Qualitative comparison under a distant single-point constraint.}
  ARDY (upper row) and Motion Mamba 2.0 (lower row) are shown at the same four timestamps. Both receive the common 18\,m target and its 33.40\,s deadline from the start. Bottom: (a) root trajectories in goal-aligned coordinates; (b) displacement along the goal direction over time. Our model makes greater progress by 25.04\,s
  (15.05 versus 9.36\,m) and has a lower final-second mean root speed (0.712 versus 1.992\,m/s).
  }
  \label{fig:distant-goal-comparison}
\end{figure*}

Although both methods receive the same distant target and deadline from the first generated window, their progress profiles differ. Motion Mamba 2.0 accumulates goal-directed displacement more continuously: it reaches 9.91 m at 16.68 s and 15.05 m at 25.04 s, compared with 6.97 m and 9.36 m for ARDY. The ARDY trajectory also contains two extended near-stationary intervals, approximately 6.7–13.6 s and 18.7–22.8 s, during which its planar root speed remains below 0.1 m/s over rolling one-second windows. It then makes a rapid late correction: its mean root speed over the final second is 1.99 m/s, compared with 0.71 m/s for Motion Mamba 2.0. This selected example suggests that Motion Mamba 2.0 allocates motion toward the distant goal earlier and more continuously, whereas ARDY defers a larger portion of its translation until later in the rollout. 
The corresponding root trajectories and quantitative
goal-progress profiles are presented in
Fig.~\ref{fig:distant-goal-comparison}.

\begin{table}[t]
\centering
\caption{\textbf{Real-Motion Reference Calibration.}
R@3 (\%) is reported for real-motion reference data
under each benchmark-specific evaluation protocol.
These values are not generated-model results or
theoretical performance bounds.}
\label{tab:reference-calibration}
\begin{tabular}{lcccc}
\toprule
Benchmark
& 2.08\,s
& 8.32\,s
& 18.72\,s
& 39.52\,s \\
\midrule
HiPHI
& 44.44
& 53.59
& 52.55
& 52.76 \\
BONES Seed
& 88.15
& 81.12
& ---
& --- \\
HumanML3D
& 91.39
& 75.96
& ---
& --- \\
\bottomrule
\end{tabular}
\end{table}

\section{Physical Evaluation Protocols}
\label{app:exp5}

\subsection{Task Completion and Tracking Measures}
\label{app:exp5-metrics}

The main suite contains 20 trials each for T1/T2 and 30 each for T3/T4. Prompt sequences are shown in Fig.~8; complete recordings are available on the project website. Let $a_{ik}$ indicate completion of stage $k$ in trial $i$, and $b_i$ indicate execution without a fall, protective stop, or human intervention. Full-task and stage success are
\begin{equation}
\begin{aligned}
\mathrm{SR}&=\frac{100}{N}\sum_i b_i\prod_{k=1}^{K_i}a_{ik},\\
\mathrm{StageSR}&=100\frac{\sum_i\sum_{k=1}^{K_i}a_{ik}}{\sum_iK_i}.
\end{aligned}
\label{eq:exp5-success}
\end{equation}
All initiated trials and planned stages enter the denominators; unreached stages fail. Standing and stopping require root speed below 0.05\,m/s for one second. A bow requires trunk flexion followed by upright recovery; circle completion requires endpoint error below 0.25\,m and heading closure within $15^\circ$. Waving, dance, and celebration are evaluated by their prescribed actions. Keyboard/mouse success requires at least 90\% of post-settling samples within 0.10\,m/s speed error and $15^\circ$ heading error, together with the final stop. Text-switch success requires action onset within 1.5\,s without a protective intervention before the next switch. Spatial-event response counts are reported separately.

For measured and commanded speed $v_t,v_t^\star$ and heading $\psi_t,\psi_t^\star$, episode-level errors are
\begin{equation}
\begin{aligned}
E_v&=\sqrt{M^{-1}\sum_t(v_t-v_t^\star)^2},\\
E_\psi&=M^{-1}\sum_t|\operatorname{wrap}(\psi_t-\psi_t^\star)|.
\end{aligned}
\end{equation}
Tracking statistics exclude the fixed 1.5-s settling allowance, which remains included in latency and deadline-response statistics. Stance slip is contact-duration-weighted horizontal foot speed in world coordinates, excluding swing phases.

\subsection{Latency and Statistical Aggregation}

Handoff latency starts upon receipt of a text or spatial command that triggers motion generation and ends when the corresponding motion reference is ready on the GPU server. It includes sampling, decoding, accepted-history updates, and text encoding when required by a text change, but excludes network communication and robot-side operations.

Physical-response latency starts at the same command receipt event and ends at the onset of a 0.5-s interval satisfying the corresponding motion tolerance. It additionally includes network transfer, controller-side buffering, processing, and physical response. Latency quantiles summarize responding events; nonresponding
or censored events remain failures in the corresponding deadline counts. Tracking and slip are aggregated as episode-level mean $\pm$ SD. Wilson intervals quantify
trial-level success rates.

\end{document}